\pdfoutput=1

\documentclass[11pt]{article}

\usepackage[preprint]{acl}

\usepackage{array}
\usepackage{enumitem}
\usepackage{times}
\usepackage{latexsym}
\usepackage{amsmath}
\usepackage{cleveref}
\usepackage{booktabs}
\usepackage{multirow}
\usepackage{pifont}
\usepackage{hyperref}       
\usepackage{url}
\usepackage{amsfonts}
\usepackage{microtype}
\usepackage{graphicx}
\usepackage{colortbl}
\usepackage{amssymb}
\usepackage{tcolorbox}
\usepackage{tabularx}     
\usepackage{caption}      
\usepackage[utf8]{inputenc}
\usepackage{longtable}

\usepackage{booktabs} 
\usepackage{pifont}   
\usepackage{booktabs}   
\usepackage{pifont}     
\usepackage{tabularx}   
\usepackage{enumitem}
\usepackage{makecell}
\usepackage{booktabs}
\usepackage{multirow}
\usepackage{xcolor}
\usepackage{amsmath}
\usepackage{graphicx}

\newcommand{\upg}[1]{\scalebox{0.75}{\textcolor{teal}{($\uparrow$#1)}}} 
\newcommand{\upb}[1]{\scalebox{0.75}{\textcolor{blue}{($\uparrow$#1)}}} 

\definecolor{Maroon}{rgb}{0.5, 0, 0}  
\definecolor{OliveGreen}{rgb}{0.33, 0.42, 0.18}  
\definecolor{Violet}{cmyk}{0.25, 0.5, 0, 0}
\usepackage[T1]{fontenc}

\usepackage[utf8]{inputenc}

\usepackage{microtype}

\usepackage{inconsolata}

\usepackage{graphicx}
\newcommand{\bcmark}{\color{Violet}{\ding{51}}}%
\newcommand{\notcheckmark}{\textcolor{black}{\bcmark\kern-1.1ex\raisebox{.7ex}{\rotatebox[origin=c]{125}{--}}}\color{black}}

\title{OmniDiagram: Advancing Unified Diagram Code Generation via Visual Interrogation Reward}

\author{
    \textbf{Haoyue Yang\textsuperscript{1,}\thanks{Equal contribution.}},
    \textbf{Xuanle Zhao\textsuperscript{1,}\footnotemark[1]},
    \textbf{Xuexin Liu\textsuperscript{1,}\footnotemark[1]}, 
    \textbf{Feibang Jiang\textsuperscript{2}},
    \textbf{Yao Zhu\textsuperscript{3,}\thanks{Corresponding author.}}
    \\
\textbf{\textsuperscript{1}} Institute of Automation, Chinese Academy of Sciences
\\
\textbf{\textsuperscript{2}} University of Chinese Academy of Sciences
\textbf{\textsuperscript{3}} Zhejiang University 
\\
\texttt{yanghaoyue2024@ia.ac.cn, ee\_zhuy@zju.edu.cn}
}

\begin{document}
\maketitle
\begin{abstract}

The paradigm of programmable diagram generation is evolving rapidly, playing a crucial role in structured visualization. However, most existing studies are confined to a narrow range of task formulations and language support, constraining their applicability to diverse diagram types. In this work, we propose OmniDiagram, a unified framework that incorporates diverse diagram code languages and task definitions. To address the challenge of aligning code logic with visual fidelity in Reinforcement Learning (RL), we introduce a novel visual feedback strategy named Visual Interrogation Verifies All (\textsc{Viva}). Unlike brittle syntax-based rules or pixel-level matching, \textsc{Viva} rewards the visual structure of rendered diagrams through a generative approach. Specifically, \textsc{Viva} actively generates targeted visual inquiries to scrutinize diagram visual fidelity and provides fine-grained feedback for optimization. This mechanism facilitates a self-evolving training process, effectively obviating the need for manually annotated ground truth code. Furthermore, we construct M3$^2$Diagram, the first large-scale diagram code generation dataset, containing over 196k high-quality instances. Experimental results confirm that the combination of SFT and our \textsc{Viva}-based RL allows OmniDiagram to establish a new state-of-the-art (SOTA) across diagram code generation benchmarks. Code, datasets, and models are available at \url{https://github.com/Haoyue-Yang/OmniDiagram}.


\end{abstract}


\section{Introduction}
Multimodal code generation has rapidly evolved beyond text-centric methods to encompass the direct manipulation and editing of visual inputs. While Large Language Models (LLMs) have significantly advanced traditional code synthesis, the rise of Multimodal Large Language Models (MLLMs) enables the processing of unstructured diagrams in raster formats (e.g., PNG), \cite{liu2024exploring, yang2024chartmimic}. This shift addresses a critical real-world demand for a unified intelligence that can bridge complex visual information with executable code across diverse tasks and languages.

\begin{figure}[t]
    \centering
    \includegraphics[width=0.45\textwidth]{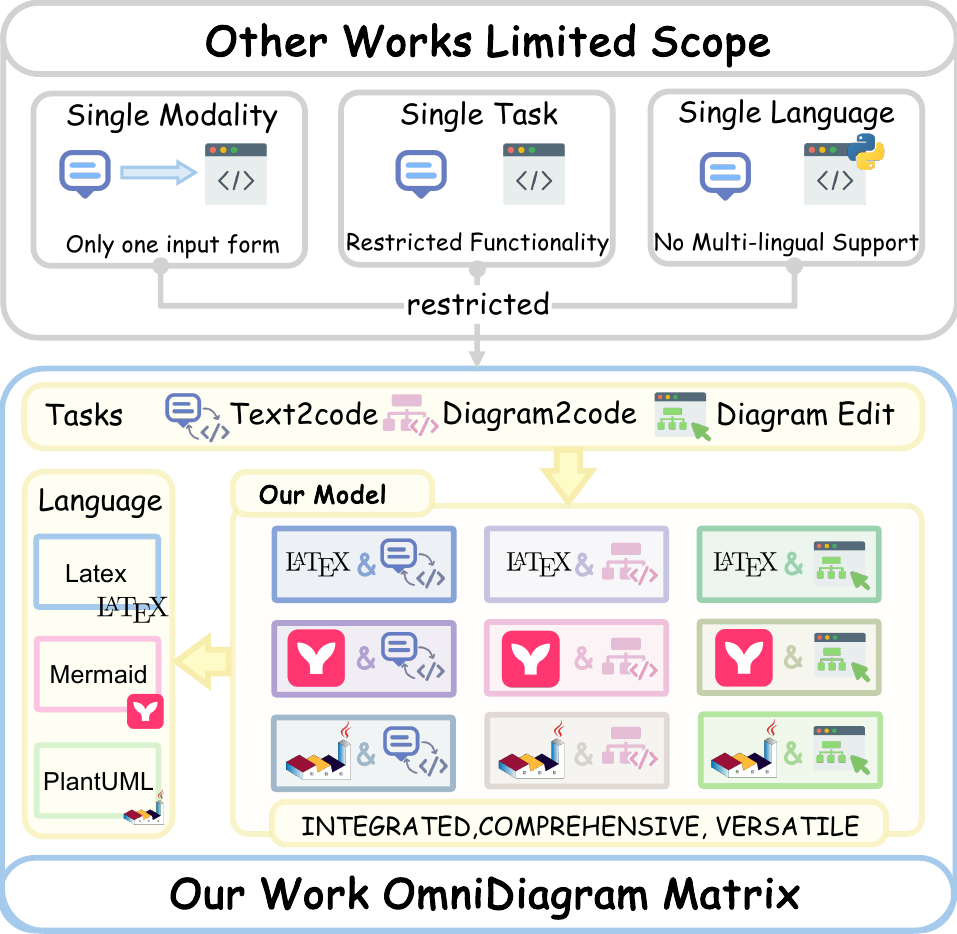}
    \vspace{-5pt}
    \caption{Overcoming the barriers of single-modality: The comprehensive landscape of OmniDiagram.}
    \label{fig:Viva_fig1}
    \vspace{-15pt}
\end{figure}

Despite meaningful explorations in diagram code generation, existing approaches generally suffer from limited versatility. They tend to be highly specialized, typically addressing singular tasks or supporting only a narrow set of programming languages. For instance, StarFlow \cite{bechard2025starflow} focuses exclusively on JSON output, neglecting diverse diagrammatic languages. Similarly, while JanusCoder \cite{sun2025januscoder} attempts to broaden task integration across Text-to-Code and Diagram-to-Code scenarios, it relies solely on Supervised Fine-Tuning (SFT). This reliance constrains the model's capacity for visual alignment and robust code execution. Other works, such as MSRL \cite{chen2025breaking} and RLRF \cite{rodriguez2025rendering}, introduce RL with a multimodal reward system. However, they remain tailored for specific image-to-code tasks, lacking the flexibility required for broader task modalities.

In this work, we focus on diagram code generation and propose OmniDiagram, a unified framework supporting diverse task modalities and code languages, as illustrated in Figure~\ref{fig:Viva_fig1}. To unify the verification of critical structural details across heterogeneous tasks and enhance the visual fidelity of the rendered diagrams, we introduce Visual Interrogation Verifies All (\textsc{Viva}).
Drawing from cognitive principles \cite{flavell1979metacognition}, \textsc{Viva} utilizes a question-driven verification strategy. Specifically, for each instance, the reward model formulates several critical questions conditioned on the instruction and input images (if available). Subsequently, the rollout code is executed and rendered into visual images, and the reward model assesses visual fidelity by answering questions conditioned on the generated images. This fine-grained, question-based feedback loop enables a continuous self-evolutionary process, allowing OmniDiagram to iteratively refine its performance across three core tasks.

To address the scarcity of diagram code generation datasets, we construct M3$^2$Diagram, the first large-scale multimodal corpus structured as a $3 \times 3$ task-language matrix. It covers \LaTeX, Mermaid, and PlantUML across Diagram-to-Code, Diagram-Editing, and Text-to-Code tasks. All instances are synthesized via a top-down pipeline to ensure code quality and visual diversity. Following rigorous filtering and verification, we curate the final dataset. To ensure robust evaluation, we cluster and partition the data into a training set and an evaluation benchmark, designated as M3$^2$Bench. Additionally, every sample in M3$^2$Bench undergoes expert human verification to guarantee correctness and reliability.
Experimental results on diagram code generation benchmarks, such as the Mermaid subset of VisPlotBench~\citep{ni2025viscoder2} and M3$^2$Bench, demonstrate the superiority of OmniDiagram, consistently surpassing both task-specific and open-source models.
In summary, our contributions are as follows:

\begin{itemize}[leftmargin=*]
    \item  We propose \textsc{Viva}, a novel generative visual feedback mechanism that serves as an instance-specific reward signal for RL. Grounded in the philosophy that every sample deserves meticulous questioning, \textsc{Viva} generates visual inquiries to evaluate the rendered execution of rollout code, thereby enhancing visual fidelity.
    
    \item We construct M3$^2$Diagram, the first large-scale omni-multimodal dataset with 196k samples, alongside M3$^2$Bench, a 1.7k verified benchmark for rigorous evaluation.

    \item  We propose OmniDiagram, the first omni diagram code generation model. Experimental results demonstrate that OmniDiagram surpasses existing open-source models significantly.
\end{itemize}

\vspace{-13pt}
\section{Related Work}
\subsection{Multimodal Code Generation} Prior research has introduced various specialized multimodal code generation methods for specific domains, such as Web2Code \cite{yun2024web2code} and ChartCoder \cite{zhao2025chartcoder}.
Recent MLLMs for code generation begin to adopt the Omni paradigm, unifying architectures for multiple code generation and editing \cite{sun2025januscoder, zhao2025vincicoder}. Specific tasks include Text-to-Code refinement \cite{rahman2025text2vis, jain2025doc2chart} and vector graphics synthesis \cite{yang2025omnisvg}. Conversely, Diagram-to-Code research employs synthetic alignment and massive datasets to enhance structural robustness in domains like UML and flowcharts \cite{bates2025unified, he2025flow2code, bechard2025starflow, chai2025multilingual, singh2024flowvqa}. For instruction-driven editing, recent benchmarks and frameworks \cite{zhao2025chartedit, chen2025charteditor} leverage rendering-aware RL to ensure execution fidelity.

\begin{figure*}[htbp]
    \centering
    \includegraphics[width=0.98\textwidth]{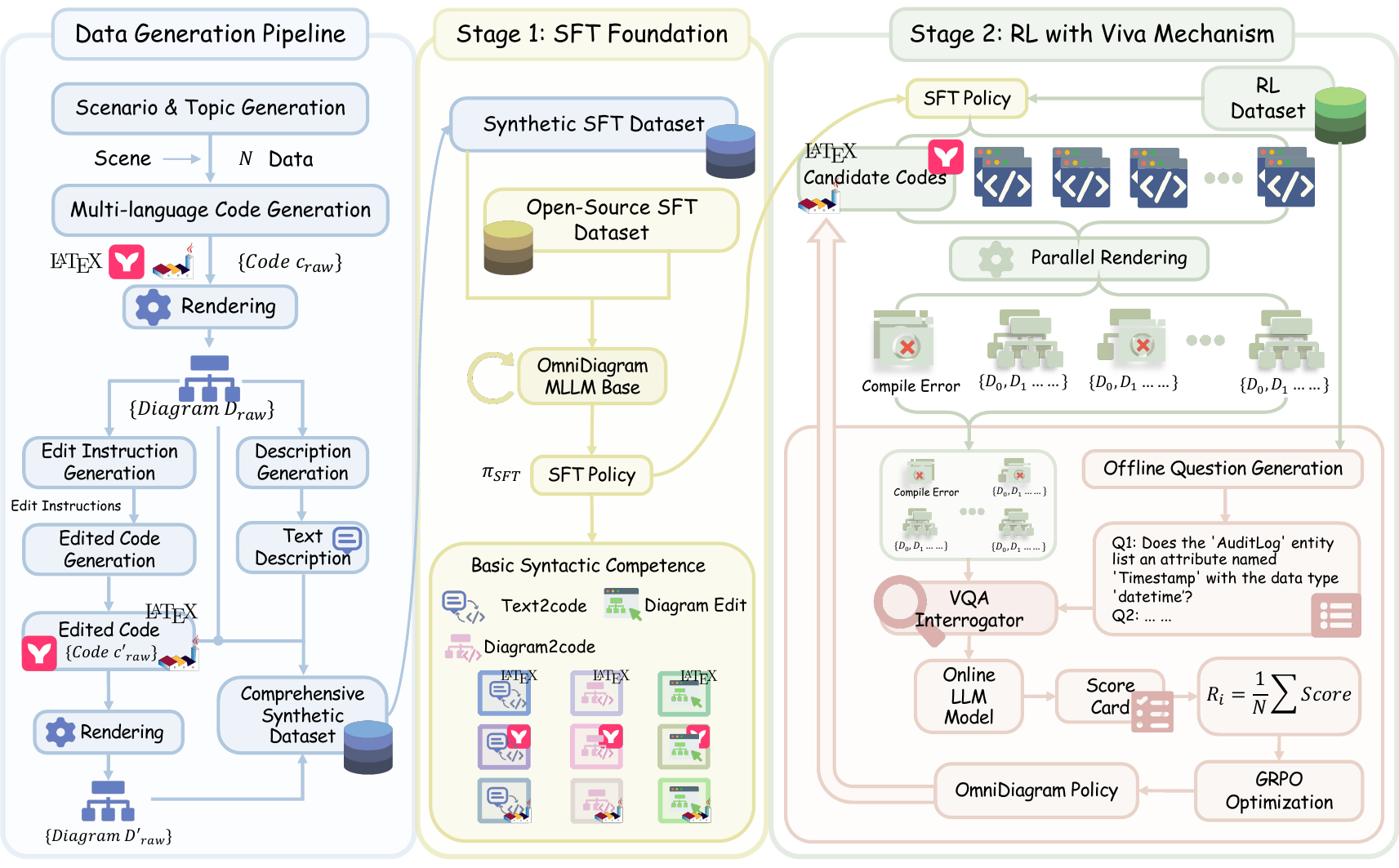}
    \vspace{-5pt}
    \caption{Overview of the OmniDiagram methodology. The framework illustrates the end-to-end flow from scalable data synthesis to model training, highlighting the two-stage optimization strategy combining an SFT foundation with the proposed \textsc{Viva}-guided RL mechanism.}
    \label{fig:Overall}
    \vspace{-13pt}
\end{figure*}

\subsection{Reinforcement Learning for Code} While Reinforcement Learning from Human Feedback (RLHF) has been pivotal for alignment, recent trends shift towards RLAIF \cite{lee2023rlaif} and environmental feedback to improve scalability. Following DeepSeek-R1 \cite{guo2025deepseek}, applying RL to enhance multimodal code generation has become a widely adopted paradigm. Specifically, to optimize reasoning paths, Step-Text2Vis \cite{luo2025nvbench} employs Step-wise Direct Preference Optimization (Step-DPO) \cite{lai2024step} for rationale refinement. In parallel, RLRF \cite{rodriguez2025rendering} proposes utilizing visual feedback with Group Relative Policy Optimization (GRPO) \cite{shao2024deepseekmath} for SVG generation. Similarly, DaVinci \cite{xingchendavinci}, MSRL \cite{chen2025breaking}, and ChartMaster \cite{tan2025chartmaster} leverage GRPO with multimodal rewards to align visual fidelity with code execution.


\section{Method}
The overall framework of OmniDiagram is illustrated in Figure~\ref{fig:Overall}. Our pipeline comprises three integral stages: (1) data generation (Section~\ref{sec:Data}), (2) Supervised Fine-Tuning (SFT) to establish foundational syntactic competence (Section~\ref{sec:SFT}), and (3) evolutionary RL driven by the \textsc{Viva} mechanism to align code with visual fidelity (Section~\ref{sec:RL}).

\subsection{Task Definition}



The Omni framework is designed to address diverse code generation tasks. In this work, we specifically focus on diagrammatic code, encompassing \LaTeX, Mermaid, and PlantUML. Let $\mathcal{I}$, $\mathcal{T}$, and $\mathcal{E}$ denote the visual diagram, textual description, and editing instruction, respectively. The model $\mathcal{M}_{\theta}$ maps an input subset $\mathcal{X}$ to the executable code $\mathcal{C}$:
\begin{equation}
    \mathcal{C} = \mathcal{M}_{\theta}(\mathcal{X}), \quad \mathcal{X} \subseteq \{\mathcal{I}, \mathcal{T}, \mathcal{E}\}
\end{equation}
This formulation encompasses three core tasks: \textit{Diagram-to-Code} ($\mathcal{X} = \{\mathcal{I}\}$), \textit{Diagram Editing} ($\mathcal{X} = \{\mathcal{I}, \mathcal{E}\}$) and \textit{Text-to-Code} ($\mathcal{X} = \{\mathcal{T}\}$), . By unifying these mappings, OmniDiagram supports both initial diagram synthesis and iterative refinement within a single architecture.

\subsection{Dataset Composition}
\label{sec:dataset_composition}
\begin{figure}[t]
    \centering
    \includegraphics[width=0.45\textwidth]{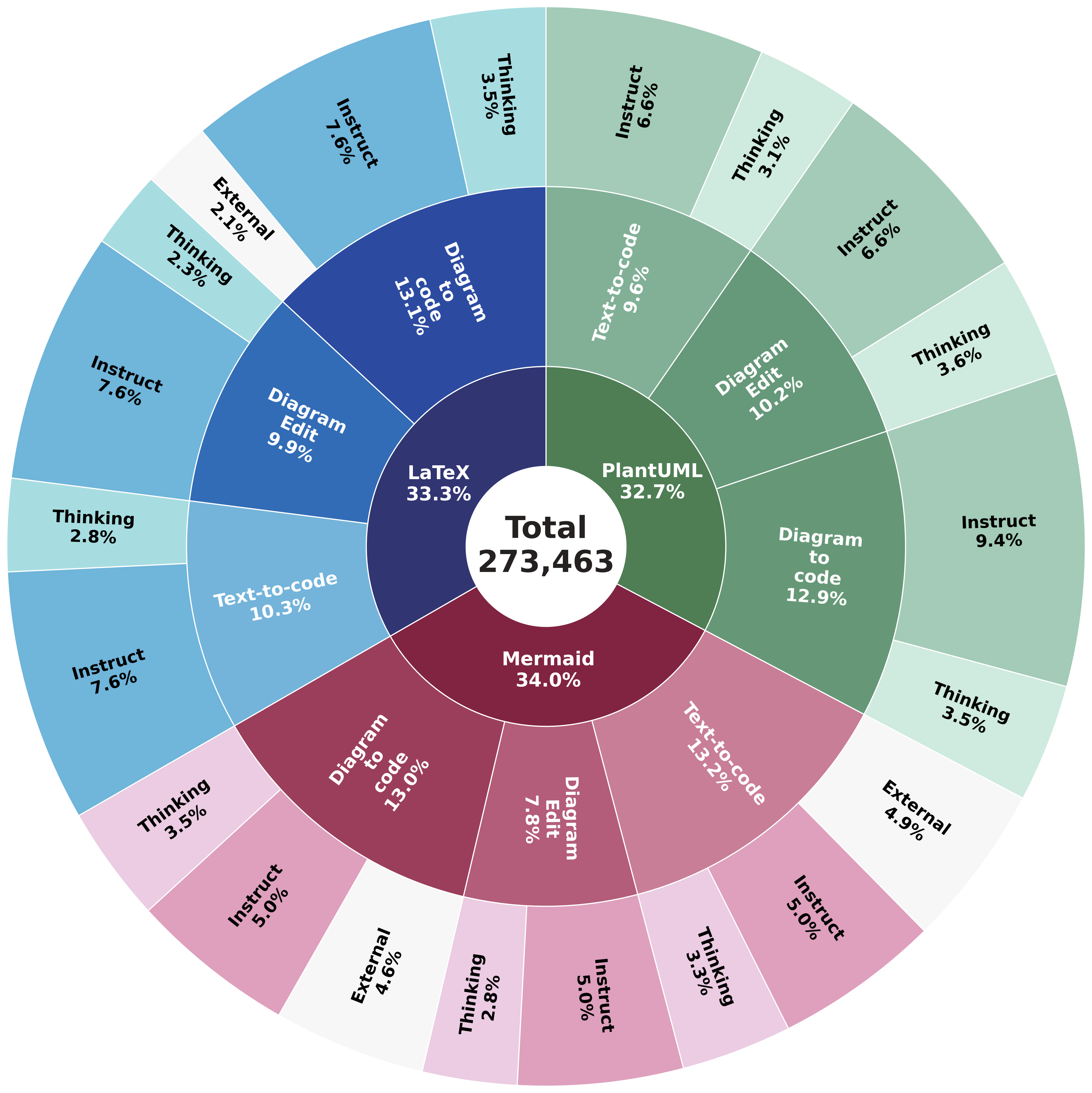}
    \vspace{-5pt}
    \caption{Breakdown of the 196k-sample M3$^2$Diagram dataset, supplemented by 77k reasoning-enriched samples, categorized across languages, tasks, and data sources.}
    \label{fig:dataset_sunburst}
    \vspace{-10pt}
\end{figure}

To address the scarcity of large-scale and diverse diagrammatic code datasets, we construct a comprehensive dataset, \textbf{M3$^2$Diagram}, comprising 196k samples for SFT and RL training. Specifically, it combines 31k open-source samples and 165k high-quality samples synthesised by us.
To ensure code correctness, we validate the alignment between code and images through rendering (tools are detailed in Appendix~\ref{app:rendering_tools}) and subsequent visual filtering. Figure~\ref{fig:dataset_sunburst} shows that the dataset maintains a balanced distribution across the three languages.

\subsubsection{Data Synthesis}
\label{sec:Data}
To address the requirements of diagrammatic code generation tasks, we construct a $3\times3$ task-language matrix encompassing \LaTeX, Mermaid, and PlantUML across three primary tasks: \textit{Diagram-to-Code}, \textit{Diagram-Editing}, and \textit{Text-to-Code}. To ensure the training data accurately reflects real-world complexity, we implement diversity constraints from the initial generation phase, requiring each language to cover approximately 15 distinct diagram types with a particular emphasis on flowchart-related variants. The detailed classification of these types is provided in Table~\ref{tab:diagram_taxonomy} of Appendix~\ref{sec:dataset_details}.

Following \cite{yang2025scaling}, we adopt a top-down, scenario-driven approach by sampling topics and scenarios to ensure semantic diversity. Leveraging Gemini-2.5-Flash, our core pipeline follows the sequence: $\texttt{topic} \rightarrow \texttt{scenario} \rightarrow \texttt{structured data} \rightarrow \texttt{code-image pairs}$. Building on the foundation of image-code pairs, the \textit{Diagram-Editing} task synthesizes editing instructions and revised code outputs, while \textit{Text-to-Code} derives corresponding textual descriptions. We employ a rigorous error-correction loop with execution feedback to ensure correctness. We employ a rigorous error-correction loop with execution feedback to ensure correctness. Additionally, rendered images are verified to eliminate non-standard outputs. This process filters an initial pool of 300k candidates into 165k high-quality, executable samples. The prompts are provided in Appendix~\ref{sec:generate_data_prompt}. 

To ensure a balanced distribution of difficulty and topological complexity of SFT and RL training, we adopt a stratified clustering strategy based on perceptual hashing. For the 196k samples, we categorize them according to their visual-structural features and partition them into SFT and RL training sets at a fixed ratio. This ensures that each subset remains uniformly distributed across different diagram types and difficulty levels, thereby mitigating structural bias in any specific partition. 
Furthermore, we extend this strategy across diverse topics to establish M3$^2$Bench, a benchmark for omni-diagram code generation. Through rigorous filtering of diagram types and visual diversity, we curate 1.7k high-quality evaluation samples designed to comprehensively assess model diagram code generation capacity. Please refer to the Appendix~\ref{app:test_data} for detailed data distribution statistics.

While recent findings from Qwen3-VL suggest that the thinking mode leads to performance degradation in multimodal code generation~\cite{bai2025qwen3vltechnicalreport}, we seek to reinvestigate its potential specifically for diagram code generation. To this end, we distill forward reasoning trajectories from Gemini-2.5-Flash and reformulate the input queries and ground-truth code to construct 77k reasoning-enriched samples. These are combined with 196k direct-generation samples to explore the impact of reasoning on diagram generation capabilities.


\subsubsection{Open-Source Data Curation}
To enhance the model's generalization capabilities, we augment our synthetic corpus with 31k samples derived from open-source datasets. Specifically, we incorporate a subset from Cosyn-400k~\cite{yang2025scaling} for the \textit{Image-to-\LaTeX} and \textit{Image-to-Mermaid} tasks. To ensure data integrity, all code snippets are standardized and validated through a rendering pipeline. For the \textit{Text-to-Mermaid} task, we integrate data from the Mermaid set of Viscoder2~\cite{ni2025viscoder2}, similarly verifying these text-code pairs before inclusion.

\subsection{Model Training}

\subsubsection{Supervised Finetuning (SFT)}
\label{sec:SFT}
Since general open-source VLMs exhibit limited proficiency in diagram code generation, we employ SFT as a preliminary stage. This establishes the fundamental diagram code generation capacity and guarantees valid candidates for subsequent refinement. The model is optimized via a standard next-token prediction objective,

\begin{small}
\begin{equation}
\mathcal{L}(\theta) = -\mathbb{E}_{(x, y)} \left[ \sum_{t=1}^T \log P(y_t \mid x, y_{<t}; \theta) \right],
\end{equation}
\end{small}
where $(x, y)$ denotes the query and target response. 

\subsubsection{\textsc{Viva} Reinforcement Learning (RL)}
\label{sec:RL}

While previous diagram code generation works rely solely on SFT for optimization, we propose utilizing RL to further enhance the executability and visual fidelity of the generated code. However, utilizing Reinforcement Learning with Verifiable Rewards (RLVR) for code generation is challenging due to verification difficulties and the lack of visual fidelity guarantees.
Current visual feedback approaches typically fall into two categories: utilizing fixed prompt templates or computing global visual similarity. The former is heavily constrained by the evaluator model's capacity and is susceptible to prompt hacking.
The latter, by over-relying on global similarity, often results in misaligned optimization objectives, biasing the model towards superficial structural resemblance at the expense of fine-grained detail accuracy.
This issue is further compounded by the unique challenges of Omni tasks: the structural diversity of \textit{Text-to-Code}, where multiple valid layouts preclude the use of a single reference image, and the non-bijective nature of \textit{Diagram-to-Code}, where visually identical outputs can stem from distinct code snippets. Inspired by the question-driven analytical inspection mechanism employed by humans when performing complex, constructive, and reasoning-intensive tasks \cite{flavell1979metacognition}, such as drawing, programming, and engineering review, we argue that effective evaluation should emulate how humans verify complex tasks—not through holistic similarity, but by systematically checking structural and semantic constraints through targeted questions.

Building upon this paradigm, we introduce \textsc{Viva}, a unified reward mechanism driven by Visual Question Answering (VQA). Departing from passive scoring, \textsc{Viva} assesses model outputs using instance-specific interrogative queries to evaluate the topological and semantic integrity of the rendered images. Specifically, for each RL training sample, we generate multiple visual questions derived from the input queries. A reward model then answers these questions based on the rendered images of the rollout code to evaluate its accuracy.

Also, to enhance the training efficiency and robustness, we decouple question generation from answer verification. We offline generate fine-grained questions per sample, specifically designed such that an answer of Yes corresponds to correct visual fidelity, as demonstrated in Appendix~\ref{sec:viva_showcase}. During training, a evaluation model interrogates each rollout candidate in an online manner. Notably, we incorporate intermediate scoring to reward partially correct outputs, yielding a smoother feedback signal that more accurately reflects the model’s performance. A detailed theoretical analysis of this reward stability is provided in Section~\ref{sec:stability_analysis}.
This approach preserves the essential benefits of visual feedback while enabling task- and language-agnostic unification across the entire Omni diagram code generation framework. By emphasizing logical consistency aligned with user intent rather than strict global imitation, \textsc{Viva} rewards more diverse rollouts during RL.


We employ GRPO for optimization \cite{shao2024deepseekmath}. For each input $x$, we generate $G$ candidate codes. These candidates are first rendered into images. Those that fail to compile are immediately assigned a reward of 0, while successfully rendered images are evaluated online by a VQA model against the pre-defined questions. Finally, candidates are ranked based on their VQA scores to compute the group-normalized advantage $\hat{A}_{i,t}$ for the $i$-th response:

\begin{small}
\begin{equation}
\hat{A}_{i,t} = \frac{R_i - \text{mean}(\{R_j\}_{j=1}^G)}{\text{std}(\{R_j\}_{j=1}^G)}.
\end{equation}
\end{small}

\noindent The policy model $\pi_{\theta}$ is optimized by maximizing the objective function, 

\vspace{-2mm}
\begin{small}
\begin{equation}
\label{eq:grpo}
\begin{aligned}
\mathcal{J}_\text{GRPO}(\theta) &= \mathbb{E}_{(x,y)\sim\mathcal{D},\substack{\{o_i\}_{i=1}^G \sim \pi_{\theta_{\text{old}}} (\cdot \mid x)}} \\
&\biggl[ \frac{1}{G} \sum_{i=1}^G \frac{1}{|o_i|} \sum_{t=1}^{|o_i|} \biggl(\min \Bigl(r_{i,t}(\theta) \hat A_{i,t}, \\
& \operatorname{clip}\left(r_{i,t}(\theta), 1-\varepsilon, 1+\varepsilon\right) \hat A_{i,t}\Bigl) \biggr) \biggr],
\end{aligned}
\end{equation}
\end{small}

\noindent where the probability ratio $r_{i,t}(\theta)$ is defined as:

\begin{small}
\begin{equation}
r_{i,t}(\theta) = \frac{\pi_\theta(o_{i,t} \mid x, o_{i,<t})}{\pi_{\theta_{\text{old}}}(o_{i,t} \mid x, o_{i,<t})}.
\end{equation}
\end{small}




\noindent To provide holistic feedback, we employ a composite reward mechanism. The final reward $R_i$ for the $i$-th candidate is defined as a weighted sum of \textsc{Viva} and format rewards,

\begin{small}
\begin{equation}
R_i = \alpha \cdot \underbrace{\left( \frac{1}{N} \sum_{k=1}^N S(\text{VQA}(o_i, q_k)) \right)}_{R_{\textsc{Viva}}} + (1-\alpha) \cdot R_{\text{fmt}}
\end{equation}
\end{small}

\noindent The \textsc{Viva} reward $R_{\textsc{Viva}}$ denotes the average \textsc{Viva} score derived from $N$ instance-specific questions. The format reward $R_{\text{fmt}} \in \{0, 1\}$ incentivizes strict adherence to the required code structures. Crucially, if the candidate code $o_i$ fails to render, $R_{\textsc{Viva}}$ is set to 0 to penalize the non-execution rollouts.

\begin{table*}[htbp]
\centering
\setlength{\tabcolsep}{1.5pt} 
\resizebox{\textwidth}{!}{ 
\begin{tabular}{l|ccc|ccc|ccc|ccc|ccc}
\toprule
\multirow{3}{*}{\textbf{Model}} & 
\multicolumn{6}{c|}{\textbf{Diagram-to-Code}} & 
\multicolumn{3}{c|}{\textbf{Diagram Editing}} & 
\multicolumn{6}{c}{\textbf{Text-to-Code}} \\
\cmidrule(lr){2-7} \cmidrule(lr){8-10} \cmidrule(lr){11-16}

 & \multicolumn{3}{c|}{\textbf{M3\boldmath$^2$Bench}} & \multicolumn{3}{c|}{\textbf{CoSyn$_\text{Diagram}$}} & 
   \multicolumn{3}{c|}{\textbf{M3\boldmath$^2$Bench}} & 
   \multicolumn{3}{c|}{\textbf{M3\boldmath$^2$Bench}} & \multicolumn{3}{c}{\textbf{VisPlot$_\text{Mermaid}$}} \\
\cmidrule(lr){2-4} \cmidrule(lr){5-7} \cmidrule(lr){8-10} \cmidrule(lr){11-13} \cmidrule(lr){14-16}

 & Exec(\%) & \boldmath$S_{vis}$ & \boldmath$S_{code}$ 
 & Exec(\%) & \boldmath$S_{vis}$ & \boldmath$S_{code}$ 
 & Exec(\%) & \boldmath$S_{pres}$ & \boldmath$S_{task}$ 
 & Exec(\%) & \boldmath$S_{vis}$ & \boldmath$S_{task}$ 
 & Exec(\%) & \boldmath$S_{vis}$ & \boldmath$S_{task}$ \\
\midrule

\multicolumn{13}{l}{\textbf{Closed-Source Models}} \\ 
GPT-5-mini & 77.6 & 62.8 & 27.0 & 76.5 & 59.8 & 21.9 & 66.6 &  52.5 & 63.5 & 87.3 & 65.1 & 81.6 & 57.5 & 34.2 & 50.5 \\
Gemini-3-Flash & 82.4 & 73.6 & 31.7 & 83.4 & 72.2 & 37.1 & 86.5 & 77.8 & 82.0 & 91.4 & 72.4 & 87.7 & 90.1 & 58.4 & 80.2 \\
\midrule
\multicolumn{13}{l}{\textbf{Open-Source VLMs}} \\
InternVL3-8B & 65.9 & 35.1 & 17.1 & 69.7 & 35.9 & 12.6 & 46.1 & 22.6 & 34.3 & 42.6 & 24.4 & 34.8 & 45.0 & 19.1 & 28.2 \\
InternVL3.5-8B & 57.4 & 30.6 & 12.0 & 73.1 & 37.6 & 15.9 & 50.9 & 21.9 & 34.7 & 43.4 & 27.4 & 37.2 & 64.9 & 34.7 & 48.4 \\
InternVL3.5-14B & 74.8 & 40.9 & 21.7 & 81.8 & 42.1 & 20.2 & 52.3 & 23.2 & 37.6 & 44.2 & 25.6 & 36.8 & 63.4 & 35.2 & 49.0 \\
Qwen3-VL-32B & 79.8 & 58.0 & 30.0 & 84.5 & 61.1 & 40.3 & 60.1 & 45.6 & 51.8 & 68.5 & 49.1 & 64.5 & 67.2 & 40.4 & 55.1 \\
InternVL3.5-38B & 57.0 & 34.3 & 21.5 & 70.9 & 42.2 & 21.9 & 49.6 & 28.1 & 40.1 & 46.6 & 29.0 & 40.2 & 63.4 & 33.8 & 45.8 \\
Qwen2.5-VL-72B & 84.6 & 55.0 & 29.0 & 89.7 & 57.3 & 34.5 & 66.9 & 36.8 & 54.0 & 68.7 & 46.9 & 61.1 & 60.3 & 31.0 & 46.0 \\

\midrule
Qwen2.5-VL-3B & 46.2 & 18.2 & 11.8 & 47.3 & 17.4 & 6.3 & 29.1 & 7.6 & 13.3 & 32.6 & 17.9 & 22.8 & 51.2 & 22.4 & 31.9 \\
\rowcolor[HTML]{D8ECE3} OmniDia-3B (SFT) & 88.6 & 69.5 & 54.1 & 87.2 & 66.1 & 54.0 & 70.1 & 54.4 & 62.2 & 86.5 & 64.0 & 81.0 & 82.4 & 45.8 & 64.3 \\
\rowcolor[HTML]{D8ECE3} OmniDia-3B (RL) & 93.0 & 72.2 & 51.6 & 93.7 & 67.1 & 57.5 & \textbf{75.5} & \textbf{59.0} & 64.8 & \textbf{90.3} & 70.8 & \textbf{85.0} & \textbf{88.6} & 49.4 & 64.5 \\
\midrule

Qwen2.5-VL-7B & 62.0 & 32.7 & 23.0 & 62.3 & 28.3 & 22.7 & 41.2 & 17.3 & 26.2 & 46.1 & 28.1 & 35.1 & 57.3 & 25.6 & 37.3 \\
\rowcolor[HTML]{D8ECE3} OmniDia-7B (SFT) & 92.9 & 74.1 & \textbf{55.6} & 89.5 & 69.9 & 50.9 & 70.1 & 54.3 & 62.1 & 88.7 & 70.6 & 83.5 & 84.0 & 45.3 & 62.6 \\
\rowcolor[HTML]{D8ECE3} OmniDia-7B (RL) & \textbf{94.3} & \textbf{75.5} & 51.3 & \textbf{96.0} & \textbf{74.5} & \textbf{60.6} & 73.3 & 57.2 & \textbf{65.2} & 89.2 & \textbf{71.8} & 84.2 & 86.3 & \textbf{51.0} & \textbf{66.9} \\
\bottomrule
\end{tabular}
}
\vspace{-5pt}
\caption{Main results across three tasks. We compare models on our proposed \textbf{M3\boldmath$^2$Bench} and existing datasets, including the diagram set of \textbf{CoSyn} and the Mermaid set of \textbf{VisPlotBench}. All metrics are 0-100. Best results are bolded among open-source models.}
\label{tab:main_results}
\vspace{-10pt}
\end{table*}

\subsection{Proof the Stability of \textsc{Viva} Rewards}
\label{sec:stability_analysis}

The effectiveness of policy gradient-based RL is profoundly influenced by the statistical variance of the reward signal. To provide theoretical insight into the stability of \textsc{Viva}, we analyze its variance properties by modeling the total reward $R_{\mathrm{acc}}$ as the average of $N$ graded QA scores $X_k \in [0,1]$. Unlike conventional binary rewards (0 or 1) that follow a high-variance Bernoulli distribution, \textsc{Viva}'s graded scoring function allows for intermediate values, which mathematically ensures that the single-item variance satisfies $\mathrm{Var}[X_k] \le \mu_k(1-\mu_k)$.

Furthermore, for analytical tractability, we model the $N$ instance-specific questions using an equicorrelation structure with correlation factor $\rho$. Under this assumption, the variance of the aggregated reward is given by:

{\begin{small}
\begin{equation}
\mathrm{Var}[R_{\mathrm{acc}}] = \frac{1}{N^2} \left( \sum_{k=1}^N \sigma_k^2 + 2 \sum_{k<l} \rho \sqrt{\sigma_k^2 \sigma_l^2} \right).
\end{equation}
\end{small}}

As derived in Appendix~\ref{app:viva_proof}, this multi-dimensional aggregation attenuates the impact of individual VQA uncertainty and provides a lower-variance advantage estimate $A_i$. Consequently, \textsc{Viva} reduces the noise propagation to the policy gradient $\hat g = \nabla_\theta \log \pi_\theta(o_i \mid x) A_i$, thereby ensuring robust convergence and smoother optimization across diverse diagrammatic tasks.

\section{Experiments}
\subsection{Implementation Details}
We separate the M3$^2$Diagram dataset into 8:1 as SFT and RL training sets.
For SFT, we fine-tune Qwen2.5-VL-3B/7B-Instruct \cite{bai2025qwen2} on the SFT dataset for two epochs on 8 H800 GPUs with a global batch size of 32. For RL, we apply GRPO to optimize the model with $G=4$ and $\alpha=0.9$. We first generate \textsc{Viva}'s queries offline via GPT-4.1-mini \cite{openai2025gpt41} and employing Qwen3-VL-32B \cite{bai2025qwen3vltechnicalreport} as the reward model during training. This stage uses a global batch size of 128 across 8 GPUs for the policy. Our implementation leverages ms-swift \cite{zhao2025swift} and  EasyR1 \cite{sheng2024hybridflow}. More details are denoted in Appendix~\ref{app:efficiency_analysis}.

\subsection{Evaluation Settings}
We evaluate the model on M3$^2$Bench and other open-source test datasets, Cosyn-400k-Diagram eval set \cite{yang2025scaling} and VisplotBench-Mermaid \cite{, ni2025viscoder2}. The M$^3$Bench comprises 1.7k samples, evenly distributed across \LaTeX{}, Mermaid, and PlantUML formats. 

We compare our model against proprietary models, including GPT-5-mini and Gemini-3-flash \cite{googledeepmind2025gemini3flash}, as well as a series of competitive open-source baselines such as the InternVL3 \cite{zhu2025internvl3} and InternVL3.5 \cite{wang2025internvl3} families (ranging from 8B to 38B), Qwen3-VL-32B \cite{bai2025qwen3vltechnicalreport}, and Qwen2.5-VL-72B \cite{bai2025qwen2}. To comprehensively assess generation quality, we employ a multi-dimensional evaluation framework leveraging GPT-4.1 as a judge to provide consistent and nuanced scoring.
Specifically, for the Diagram-to-Code task, we decouple visual fidelity ($S_{vis}$) from code accuracy ($S_{code}$), with the latter assessed using CrystalBLEU \cite{eghbali2022crystalbleu}. For the Diagram Editing task, we concurrently evaluate content preservation ($S_{pres}$) and instruction adherence ($S_{task}$). Regarding the Text-to-Code task, we measure visual correctness ($S_{vis}$) and task adherence ($S_{task}$), following the evaluation methodology defined in \cite{ni2025viscoder2}. Detailed scoring criteria and prompts are provided in Appendix~\ref{app:metrics} and Appendix~\ref{app:judge}, respectively.

\subsection{Main Results}

\begin{table}[t]
\centering
\small
\setlength{\tabcolsep}{2.5pt} 
\begin{tabular}{@{}llccc@{}}
\toprule
\textbf{Data} & \textbf{Model} & \textbf{Exec (\%)} & \textbf{$S_{vis}$} & \textbf{$S_{task}$} \\ \midrule

\multirow{10}{*}{\rotatebox{90}{\textbf{M3$^2$Bench}}} 
& \textit{Qwen2.5-Coder-3B} & 66.4 & 40.0 & 46.3 \\
& Viscoder2-3B & 77.9 \upb{11.5} & 54.3 \upb{14.3} & 70.5 \upb{24.2} \\
\cmidrule(lr){2-5}
& \textit{Qwen2.5-VL-3B} & 32.6 & 17.9 & 22.8 \\
& OmniDia-3B (SFT) & 86.5 \upg{53.9} & 64.0 \upg{46.1} & 81.0 \upg{58.2} \\
& OmniDia-3B (RL) & 90.3 \upg{57.7} & 70.8 \upg{52.9} & 85.0 \upg{62.2} \\
\cmidrule(lr){2-5}
& \textit{Qwen2.5-Coder-7B} & 73.0 & 50.2 & 63.3 \\
& Viscoder2-7B & 83.6 \upb{10.6} & 63.0 \upb{12.8} & 78.8 \upb{15.5} \\
\cmidrule(lr){2-5}
& \textit{Qwen2.5-VL-7B} & 46.1 & 28.1 & 35.1 \\
& OmniDia-7B (SFT) & 88.7 \upg{42.6} & 70.6 \upg{42.5} & 83.5 \upg{48.4} \\
& OmniDia-7B (RL) & 89.2 \upg{43.1} & 71.8 \upg{43.7} & 84.2 \upg{49.1} \\ \midrule

\multirow{12}{*}{\rotatebox{90}{\textbf{VisPlotBench}}} 
& \textit{Qwen2.5-Coder-3B} & 74.1 & 30.0 & 38.0 \\
& Viscoder2-3B & 76.3 \upb{2.1} & 43.0 \upb{13} & 59.0 \upb{21} \\
\cmidrule(lr){2-5}
& \textit{Qwen2.5-VL-3B} & 51.2 & 22.4 & 31.9 \\
& OmniDia-3B (SFT) & 82.4 \upg{31.2} & 45.8 \upg{23.4} & 64.3 \upg{32.4} \\
& OmniDia-3B (RL) & 88.6 \upg{37.4} & 49.4 \upg{27.0} & 64.5 \upg{32.6} \\
\cmidrule(lr){2-5}
& \textit{Qwen2.5-Coder-7B} & 77.9 & 39.0 & 53.0 \\
& Viscoder2-7B & 78.6 \upb{0.7} & 43.0 \upb{4.0} & 59.0 \upb{6.0} \\
\cmidrule(lr){2-5}
& \textit{Qwen2.5-VL-7B} & 57.3 & 25.6 & 37.3 \\
& OmniDia-7B (SFT) & 84.0 \upg{26.7} & 45.3 \upg{19.7} & 62.6 \upg{25.3} \\
& OmniDia-7B (RL) & 86.3 \upg{29.0} & 51.0 \upg{25.4} & 66.9 \upg{29.6} \\ \bottomrule
\end{tabular}
\vspace{-5pt}
\caption{Comparison on Text-to-Code tasks across M3$^2$Bench and VisPlotBench. \upb{x} (blue) denotes improvement of Viscoder2 over its Qwen-Coder baseline, while \upg{x} (green) denotes OmniDia's improvement over its Qwen-VL baseline.}
\label{tab:viscoder_vertical}
\vspace{-10pt}
\end{table}

\begin{table*}[t]
\centering
\resizebox{\textwidth}{!}{
\begin{tabular}{l cc cc cc}
\toprule
\multirow{2}{*}{\textbf{Configuration Strategy}} & \multicolumn{2}{c}{\textbf{Diagram-to-Code}} & \multicolumn{2}{c}{\textbf{Diagram Editing}} & \multicolumn{2}{c}{\textbf{Text-to-Code}} \\
\cmidrule(lr){2-3} \cmidrule(lr){4-5} \cmidrule(lr){6-7}
& \textbf{Exec (\%)} & $\overline{\mathbf{S}}_{D}$ & \textbf{Exec (\%)} & $\overline{\mathbf{S}}_{E}$ & \textbf{Exec (\%)} & $\overline{\mathbf{S}}_{T}$ \\
\midrule
\rowcolor[HTML]{F2F2F2} \multicolumn{7}{l}{\textit{Exp 1: Impact of Reasoning Trajectories}} \\
SFT (w/o Reasoning Data)           & 88.6 & 61.8 & 70.1 & 58.3 & 86.5 & 72.5 \\
SFT + RL (w/o Reasoning Data)      & 93.0 & 61.9 & 75.5 & 61.9 & 90.3 & 77.9 \\
SFT (w/ Reasoning Data)           & 81.3 & 51.5 & 71.2 & 58.7 & 68.0 & 57.8 \\
SFT + RL (w/ Reasoning Data)      & 89.3 & 55.4 & 77.6 & 62.6 & 88.1 & 76.0 \\
\midrule
\rowcolor[HTML]{F2F2F2} \multicolumn{7}{l}{\textit{Exp 2: Training Pipeline Strategy}} \\
Pure RL (w/o SFT Stage)         &  30.2 & 19.6   & 28.7 & 8.9   & 34.1   & 25.4   \\
Pure SFT (w/o RL Stage)             & 88.6 & 61.8 & 70.1 & 58.3 & 86.5 & 72.5 \\
Full Pipeline          & 93.0 & 61.9 & 75.5 & 61.9 & 90.3 & 77.9 \\
\midrule
\rowcolor[HTML]{F2F2F2} \multicolumn{7}{l}{\textit{Exp 3: Reward Model Scale}} \\
Reward by Qwen3-VL-32B              & 93.0 & 61.9 & 75.5 & 61.9 & 90.3 & 77.9 \\
Reward by Qwen3-VL-30B-A3B          & 92.4 & 60.4 & 74.5 & 64.7 & 89.2 & 76.2   \\
\bottomrule
\end{tabular}
}
\vspace{-5pt}
\caption{Comprehensive ablation studies. We report the execution rate (\textbf{Exec (\%)}) and mean scores ($\overline{\mathbf{S}}$) for each specific task category: Diagram-to-Code, Diagram Editing, and Text-to-Code.}
\label{tab:unified_ablation_final}
\vspace{-10pt}
\end{table*}

As shown in Table~\ref{tab:main_results}, OmniDiagram demonstrates leading performance in open-source logic diagram generation by consistently surpassing competitive baselines like the Qwen2.5-VL-72B across all tasks. Post-RL, OmniDiagram achieves performance comparable to, or in certain domains exceeding, proprietary models. These results demonstrate the efficacy of our unified framework in capturing the intricate structural and semantic nuances across diverse logic diagram formats. All the results validate the efficiency of our M3$^2$Diagram dataset and the \textsc{Viva} training strategy.

\begin{figure}[t]
    \centering
    \includegraphics[width=0.45\textwidth]{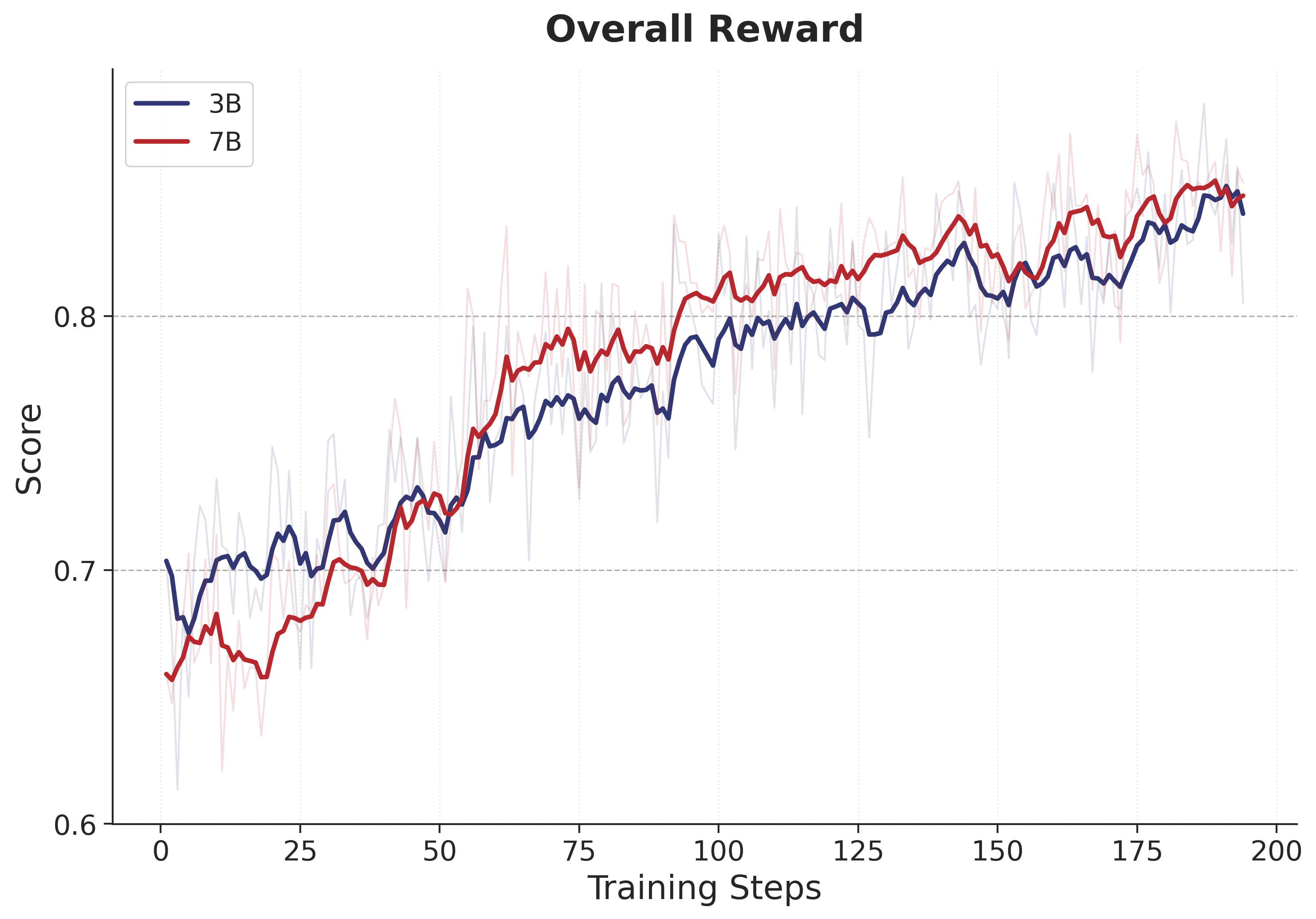}
    \vspace{-5pt}
    \caption{Progression of the overall reward during the RL training process for OmniDiagram.}
    \label{fig:rl_reward}
    \vspace{-10pt}
\end{figure}

The comparison between the SFT baseline and the final RL-enhanced models further validates our proposed two-stage training paradigm. While the SFT stage provides a robust foundation for multi-format diagram generation, the subsequent stage with \textsc{Viva} mechanism delivers substantial improvements. Specifically, the execution rate boosts significantly as the RL process penalizes non-renderable rollouts, effectively pruning invalid syntax. Concurrently, the \textsc{Viva}-based visual feedback loop directly optimizes visual perceptual alignment, resulting in significant improvements in metrics such as $S_{\text{vis}}$ and $S_{\text{fid}}$. As illustrated in Figure~\ref{fig:rl_reward}, the steady progression of the overall reward for both 3B and 7B models throughout the training process confirms the robustness of our RL strategy. This trend demonstrates that the framework effectively translates fine-grained visual feedback into superior generative performance.

We further compare the Text-to-Code performance of OmniDiagram against specialized code generation VLMs such as VisCoder2 \cite{ni2025viscoder2}. We only utilize the Mermaid set of M3$^2$Bench and VisplotBench for fair comparison.
This comparison is particularly challenging as VisCoder2 and its baselines are built upon dedicated code-centric LLM (Qwen2.5-Coder), whereas OmniDiagram originates from a general-purpose multimodal backbone  (Qwen2.5-VL). As shown in Table~\ref{tab:viscoder_vertical}, despite starting from a general-purpose backbone, OmniDiagram achieves greater performance gains than Viscoder2 relative to their respective baselines. For instance, at the 7B scale of M3$^2$Bench, our model achieves a 43.1\% leap in execution rate, far exceeding the 10.6\% improvement of VisCoder2-7B. The results demonstrate the strong capability of the unified framework and \textsc{Viva} mechanism in optimizing multimodal models for diagrammatic code synthesis.

Due to page limits, we provide a gallery of generated samples in Appendix~\ref{examples_of_ability}. 
We also conduct a systematic failure mode analysis in Appendix~\ref{app:failure_analysis} to investigate the model's limitations.

\subsection{Ablation Studies}

To verify the effectiveness of our design choices, we conduct three systematic ablation experiments focusing on the 3B model variant. All results are summarized in Table~\ref{tab:unified_ablation_final}, representing the macro-average performance across all nine tasks.

\noindent \textbf{Impact of Reasoning Trajectories.} 
The integration of reasoning data introduces a task-specific performance bifurcation as shown in Table~\ref{tab:unified_ablation_final}. Reasoning trajectories significantly bolster Diagram Editing performance, indicating that the reasoning process enhances the model instruction analysis capability. For other tasks, the extended context required for explicit reasoning may distract the model's attention from salient information, potentially leading to a degradation in performance. Detailed analysis is provided in Appendix~\ref{app:ablation_analysis}. This phenomenon could also be validated in Qwen3-VL \cite{bai2025qwen3vltechnicalreport}, where the performance of the instruct version surpasses the thinking version on multimodal code generation benchmarks.

\noindent \textbf{Training Pipeline Strategy.} 
We also conduct ablation studies on the training pipelines to validate the necessity of the SFT-to-RL strategy. As shown in Table~\ref{tab:unified_ablation_final}, the RL-Only variant fails to converge effectively on omni-diagrammatic code tasks. Detailed analysis reveals that, in the absence of SFT, the model tends toward mode collapse, exclusively generating Mermaid code while ignoring the specific requirements provided in the instructions. This behavior results in negligible execution rates for the \LaTeX{} and PlantUML test sets. We attribute this to the fact that models without SFT lack the capacity for rigorous instruction following.
In contrast, our full pipeline (\textit{w/ SFT + RL}) leverages SFT to establish foundational knowledge and RL to enforce logical discipline, achieving the best overall performance, which demonstrates that the SFT stage is essential for establishing a foundational omni-diagrammatic generation capability.

\noindent \textbf{Robustness to Reward Model Scale.} We evaluate the robustness of \textsc{Viva} by employing Qwen3-VL-32B and Qwen3-VL-30B-A3B as reward models. As demonstrated in Table~\ref{tab:unified_ablation_final}, the marginal performance discrepancy across these scales highlights the inherent robustness of our mechanism, suggesting it is relatively invariant to the capacity of the reward model.
This suggests that the offline-generated visual questions, which provide essential visual grounding, are more pivotal to the optimization than the scale of the online feedback. By structuring the visual focus beforehand, \textsc{Viva} lowers the discriminative burden on the reward model, enabling high-fidelity even with small verifiers.

\section{Conclusion}
In this work, we address the increasing demand for versatility in multimodal diagram code generation through three primary contributions. First, we introduce M3$^2$Diagram-196k, the first large-scale instruction-tuning dataset covering a $3\times3$ task-language matrix. Second, we propose \textsc{Viva}, a VQA-based reward mechanism built on the philosophy that ``every sample deserves meticulous interrogation'' via visual feedback. Third, we introduce OmniDiagram, an omni-model family that consistently surpasses competitive open-source baselines on M3$^2$Bench and various external diagrammatic benchmarks. We believe that both the M3$^2$Diagram dataset and the \textsc{Viva} framework will significantly catalyze further developments in the field of unified multimodal code generation.

\section*{Limitation}
From our perspective, our work has several limitations:
(1) Reward Weighting: The weighting $\alpha$ between visual and format rewards is currently fixed. Exploring task-specific or dynamic adjustments could further optimize performance for complex diagram types.
(2) Algorithmic Diversity: We primarily utilize GRPO for its efficiency. Comparing various RL paradigms like PPO or DPO within the \textsc{Viva} framework would provide deeper insights into different optimization strategies.
(3) Computational Overhead: The reliance on external models for data synthesis and evaluation involves significant resource investment. Future work could focus on more cost-effective methods to improve accessibility.

\section*{Ethic Statement}
Our research utilizes publicly available open-source models and datasets, all of which are properly cited. By leveraging these widely recognized and vetted resources, we significantly mitigate the risk of generating harmful or toxic content. The diagram generation tasks focus on structured visualization and technical data, ensuring the outputs remain within professional and non-sensitive domains.

\bibliography{custom}

\clearpage
\appendix

\section{Dataset Details}

\subsection{Diagram Type Statistics} 
\label{sec:dataset_details}
In this section, we present a quantitative breakdown of the diagram types generated during the Data Synthesis phase, detailing the proportional distribution of each category across the three supported syntaxes. As illustrated in Table~\ref{tab:diagram_taxonomy}, the dataset encompasses a diverse taxonomy: 12 types for \LaTeX, 13 for Mermaid, and 20 for PlantUML. This extensive coverage is predominantly centered around flowcharts and logic-driven diagrams, ensuring the dataset's robustness and applicability to complex real-world scenarios.

\begin{table*}[htbp]
\centering
\resizebox{\textwidth}{!}{%
\begin{tabular}{lclclc}
\toprule
\multicolumn{2}{c}{\textbf{\LaTeX}} & \multicolumn{2}{c}{\textbf{Mermaid}} & \multicolumn{2}{c}{\textbf{PlantUML}} \\
\cmidrule(lr){1-2} \cmidrule(lr){3-4} \cmidrule(lr){5-6}
\textbf{Diagram Type} & \textbf{Ratio (\%)} & \textbf{Diagram Type} & \textbf{Ratio (\%)} & \textbf{Diagram Type} & \textbf{Ratio (\%)} \\ 
\midrule
Graph (Nodes \& Edges) & 28.6 & Flowchart & 12.9 & Activity Diagram & 45.9 \\
Flowchart & 13.8 & Class Diagram & 12.3 & Sequence Diagram & 23.1 \\
Block Diagram & 13.7 & Kanban Board & 10.6 & Use Case Diagram & 13.5 \\
Timeline & 10.0 & Timeline & 9.6 & State Diagram & 6.6 \\
Finite State Machine & 7.5 & Entity Relationship & 9.4 & Deployment Diagram & 3.5 \\
Tree Diagram & 6.5 & Sequence Diagram & 8.3 & Regex Diagram & 2.7 \\
Unknown / Generic & 5.7 & Packet Diagram & 7.7 & Class Diagram & 2.1 \\
Chemical Structure & 4.7 & User Journey & 7.3 & Component Diagram & 1.2 \\
Free Body Diagram & 3.2 & Block Diagram & 6.6 & EBNF Diagram & 0.9 \\
Circuit Diagram & 3.1 & C4 Architecture & 5.7 & Network Diagram & 0.4 \\
Vector Graphics & 1.6 & GitGraph & 5.0 & \textit{Others**} & 0.1 \\
\textit{Others (Plots, Tables, etc.)} & 1.6 &\textit{Others* } & 4.6 & & \\
\bottomrule
\end{tabular}%
}
\vspace{1mm}
\raggedright
\footnotesize{\textit{*Others for Mermaid include Mindmaps, Gantt, State, and BPMN diagrams.}\\
\textit{**Others for PlantUML include Entity-Relationship, SDL, Timing, Wireframe, and Ditaa diagrams.}}
\vspace{2mm}
\caption{Distribution of diagram types within the dataset. The table presents the percentage breakdown of each schema for \LaTeX, Mermaid, and PlantUML. This demonstrates the dataset's coverage of diverse structural complexities ranging from scientific plots to software engineering diagrams.}
\label{tab:diagram_taxonomy}
\end{table*}

\subsection{Prompt Used for Data Generation}
\label{sec:generate_data_prompt}

In this section, we provide the detailed prompt templates employed in our top-down data generation pipeline, taking the \textit{Diagram-to-Mermaid-Mindmap} task as a representative example. To foster reproducibility and transparency, we present the prompts sequentially across the four key stages of our synthesis workflow: (1) Topic generation Figure~\ref{fig:prompt_topic}, (2) Scenario design Figure~\ref{fig:prompt_scenario}, (3) structured Data synthesis (e.g., node and edge definitions) Figure~\ref{fig:prompt_data}, and (4) the final Code implementation Figure~\ref{fig:prompt_code}. Collectively, these prompts ensure both the semantic diversity and syntactic correctness of the OmniDiagram dataset.

\subsection{Rendering Tools and Pipelines}
\label{app:rendering_tools}
To ensure high-fidelity visual outputs, we employ standardized rendering pipelines for each diagrammatic language. 
For \textbf{\LaTeX{}}, we utilize the TeX Live distribution, compiling code via \texttt{pdflatex} and converting the output to high-resolution images using \texttt{Poppler} utilities. 
For \textbf{Mermaid}, we adopt the official \texttt{mermaid-cli} based on a headless Chromium browser to guarantee browser-consistent rendering. 
Finally, for \textbf{PlantUML}, we use the standard Java engine integrated with \textbf{Graphviz} to accurately compute complex node layouts and generate anti-aliased diagrams.

\subsection{M3$^2$Bench Test Set Distribution}
\label{app:test_data}
\begin{figure}[t]
    \centering
    \includegraphics[width=0.45\textwidth]{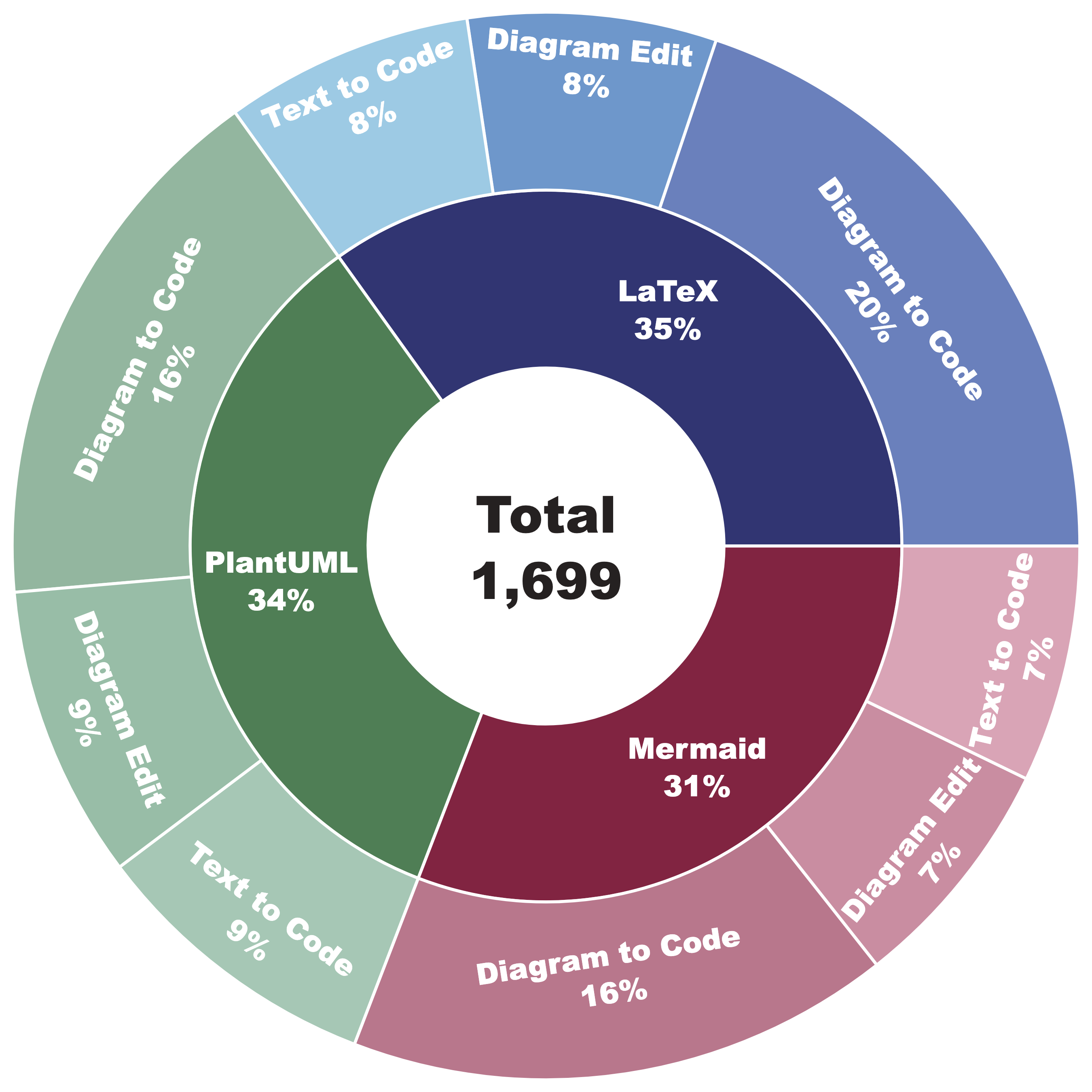}
    \vspace{-5pt}
    \caption{Statistical breakdown of tasks and diagram formats in M3$^2$Bench}
    \label{fig:viva_donut}
    \vspace{-13pt}
\end{figure}

The M3$^2$Bench test set comprises a total of 17k high-quality samples, ensuring a rigorous evaluation of model performance across diverse diagramming languages and tasks. Specifically, the test set is distributed across three primary formats: LaTeX with 593 samples, Mermaid with 524 samples, and PlantUML with 582 samples. Each language category is further divided into three functional tasks to assess versatility: Diagram-to-Code (including 337 LaTeX, 280 Mermaid, and 280 PlantUML samples), Diagram Edit (comprising 128 LaTeX, 122 Mermaid, and 151 PlantUML samples), and Text-to-Diagram (consisting of 128 LaTeX, 122 Mermaid, and 151 PlantUML samples). As illustrated in Figure~\ref{fig:viva_donut}.

\begin{figure*}[t]
    \centering
    \includegraphics[width=1.0\textwidth]{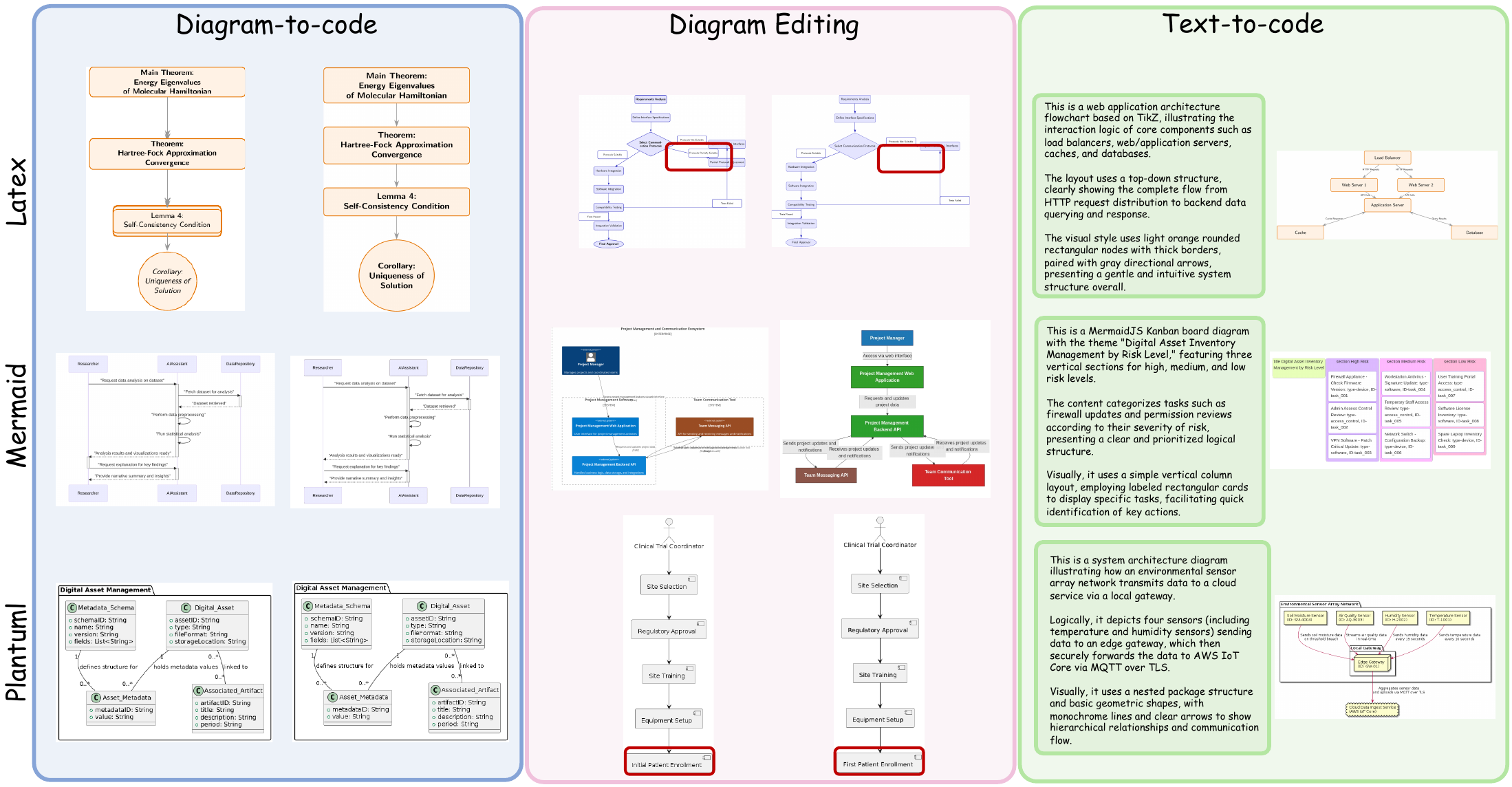}
    \vspace{-5pt}
    \caption{Qualitative showcase of our model across three modalities (\LaTeX{}, Mermaid, PlantUML) and three tasks. \textbf{Left:} Diagram-to-Code reconstruction showing high semantic accuracy. \textbf{Middle:} Diagram Editing demonstrating precise local updates (highlighted in red). \textbf{Right:} Text-to-Code generation converting complex natural language descriptions into valid structural code.}
    \label{fig:viva_showcase}
    \vspace{-13pt}
\end{figure*}

\section{Visual Verification Showcase}
\label{sec:viva_showcase}

To further illustrate the operational details of the \textsc{Viva} reward mechanism, we present a series of qualitative examples across our three primary tasks. As discussed in the main text, \textsc{Viva} deviates from traditional pixel-wise comparison by employing instance-specific "interrogative" probes. For every sample in the evaluation set, we generate ten fine-grained questions that scrutinize the rendered output from multiple dimensions: \textbf{topological structure} (connection logic), \textbf{semantic consistency} (textual accuracy), and \textbf{aesthetic attributes} (style, color, and shapes).

\begin{figure*}[t]
    \centering
    \includegraphics[width=1.0\textwidth]{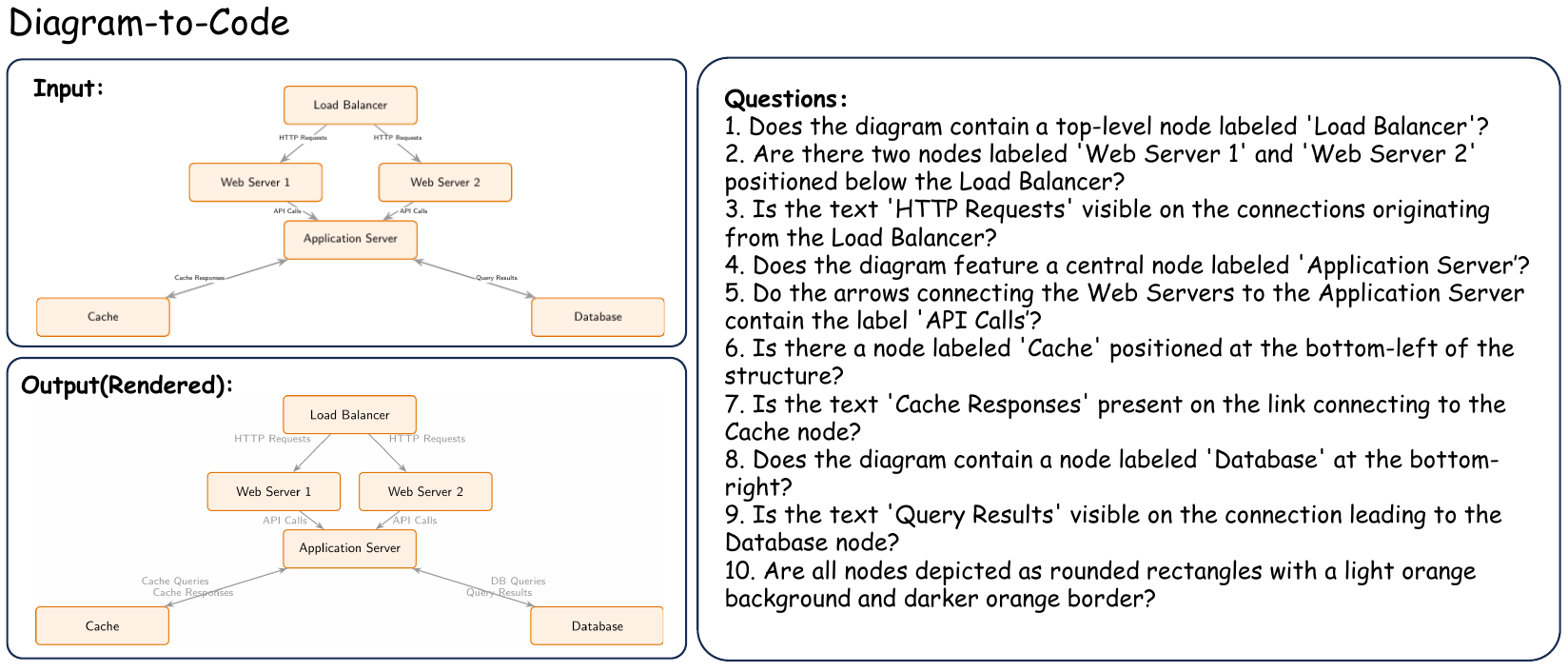}
    \vspace{-5pt}
    \caption{Qualitative example of visual verification questions for the Diagram-to-Code task.}
    \label{fig:Q_showcase_D2C}
    \vspace{-13pt}
\end{figure*}

\begin{figure*}[t]
    \centering
    \includegraphics[width=1.0\textwidth]{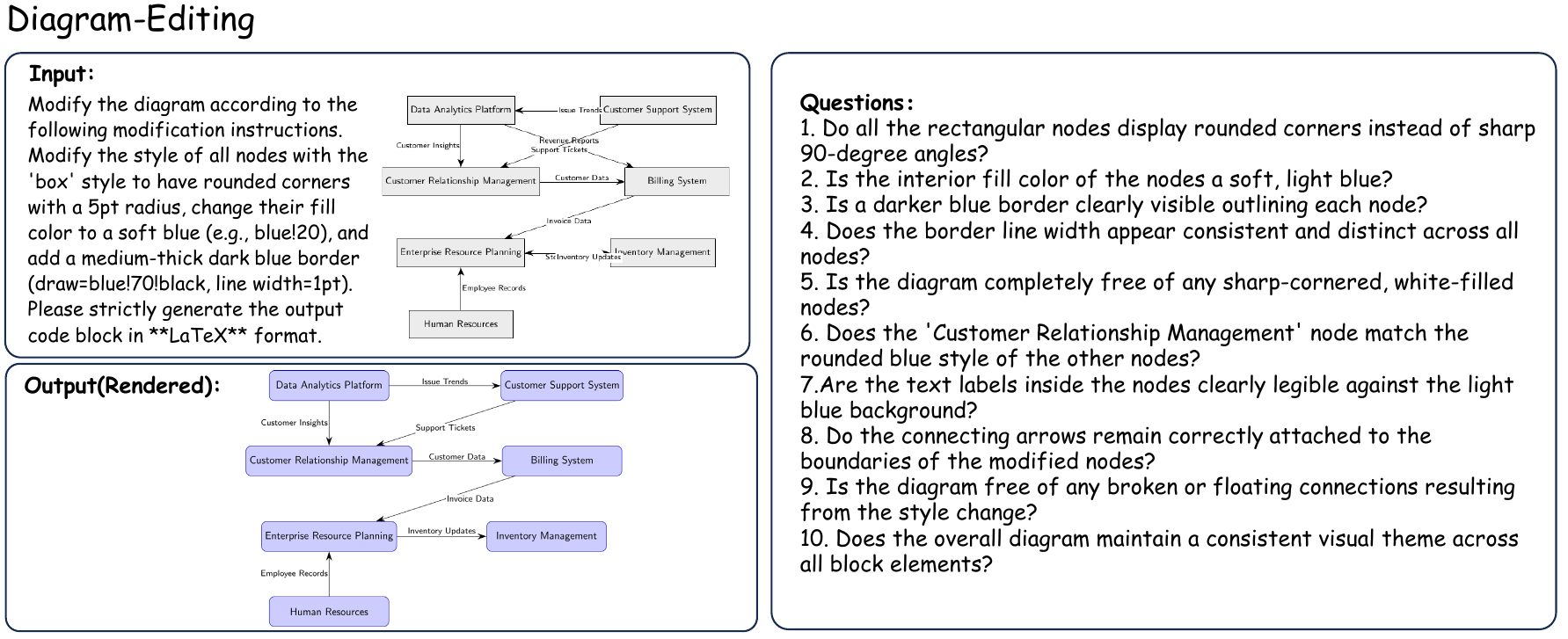}
    \vspace{-5pt}
    \caption{Qualitative example of visual verification questions for the Diagram Editing task.}
    \label{fig:Q_showcase_DE}
    \vspace{-13pt}
\end{figure*}

Figures~\ref{fig:Q_showcase_T2C}, \ref{fig:Q_showcase_D2C}, and \ref{fig:Q_showcase_DE} demonstrate the diversity and specificity of these verification probes:

\begin{figure*}[t]
    \centering
    \includegraphics[width=1.0\textwidth]{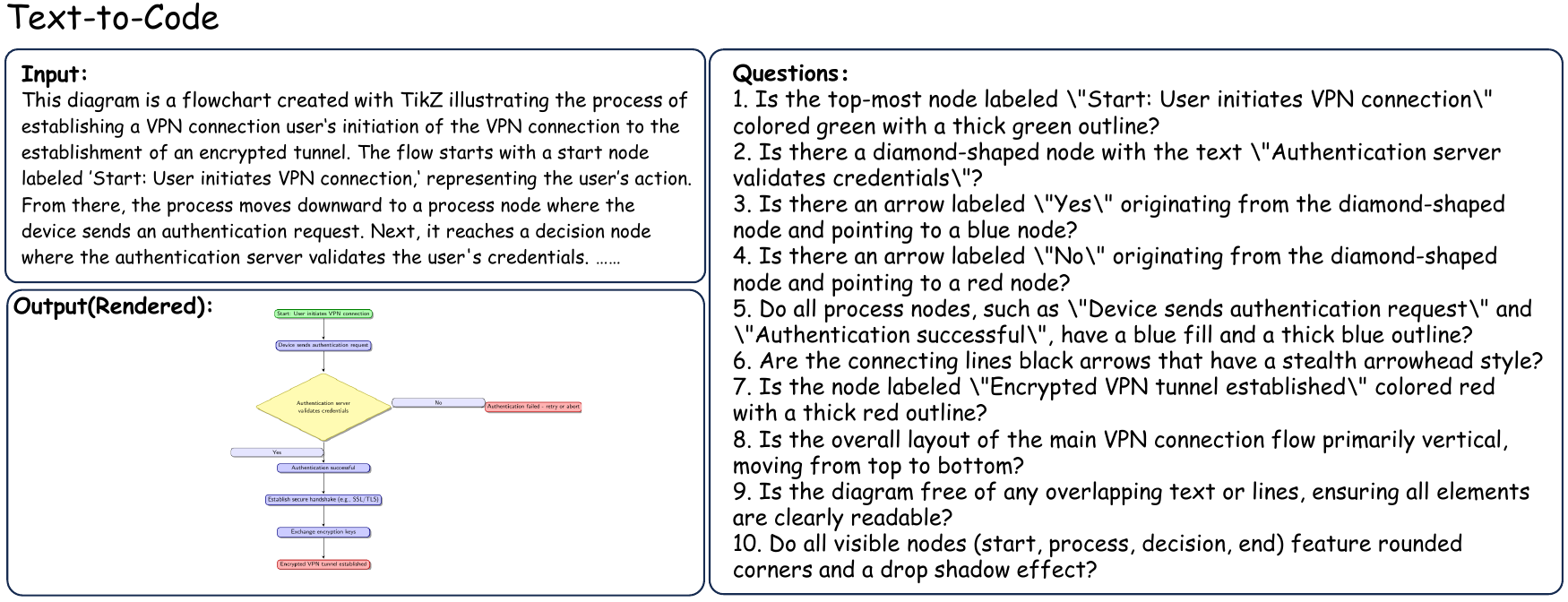}
    \vspace{-5pt}
    \caption{Qualitative example of visual verification questions for the Text-to-Code task.}
    \label{fig:Q_showcase_T2C}
    \vspace{-13pt}
\end{figure*}

In the \textbf{Text-to-Code} task, Figure~\ref{fig:Q_showcase_T2C}, the questions are designed to verify if the model has successfully translated complex textual descriptions into precise visual logic, such as identifying the correct shape for decision nodes or the specific color of a thick green outline. For the \textbf{Diagram-to-Code} task, Figure~\ref{fig:Q_showcase_D2C}, the probes focus on structural fidelity, ensuring that hierarchical dependencies (e.g., the position of 'Web Servers' relative to the 'Load Balancer') and connection labels like 'API Calls' are preserved during the code generation and rendering process. 

Finally, in the \textbf{Diagram Editing} task, Figure~\ref{fig:Q_showcase_DE}, \textsc{Viva} serves as a safeguard for instruction following. It checks whether requested modifications—such as rounding rectangle corners or applying a soft blue fill—have been applied consistently while maintaining the integrity of the original graph's connectivity. By standardizing the ground-truth response to "Yes" for all probes, \textsc{Viva} provides a task-agnostic scoring metric that facilitates stable Reinforcement Learning. Furthermore, the use of intermediate scoring ensures that models receiving partial credit for correctly rendered sub-components are appropriately rewarded, leading to a smoother and more informative feedback signal.

\section{Detailed Theoretical Proof of Reward Stability}
\label{app:viva_proof}

In policy gradient-based RL, the stability of gradient estimates largely depends on the statistical variance of the reward signal. For generative tasks with complex semantic and structural constraints, highly noisy rewards can significantly affect training stability and convergence. \textsc{Viva} RL employs a graded reward design based on multiple QA constraints. This section analyzes the variance properties of this reward mechanism and its implications for policy optimization stability.

\noindent \textbf{Modeling Graded QA Rewards as Random Variables:} For a given input $x$ and its candidate generation $o_i$, we construct $N$ QA questions $\{q_k\}_{k=1}^N$ targeting topological and semantic constraints. The score of the $k$-th QA is defined as a random variable:
$
X_k \triangleq S(\mathrm{VQA}(o_i, q_k)) \in [0,1],
$
where $S(\cdot)$ is a graded scoring function that allows intermediate values (e.g., 0.5) to represent partial semantic satisfaction. The randomness mainly arises from visual rendering errors, VQA inference uncertainty, and ambiguity in semantic judgment. Moreover, we allow each QA to have its own mean and variance:
$
\mathbb{E}[X_k] = \mu_k, \mathrm{Var}[X_k] = \sigma_k^2.
$

To capture the correlation induced by shared underlying semantic structure, we assume an equicorrelation structure for analytical convenience:

\begin{small}
\begin{equation}
\mathrm{Cov}[X_k, X_l] = \rho \sqrt{\sigma_k^2 \sigma_l^2}, \quad k \neq l, \quad \rho \in [0,1),
\end{equation}
\end{small}

where $\rho$ measures the degree of dependency between QA scores.

\noindent \textbf{Variance of Averaged Graded QA Rewards:} The accuracy reward in \textsc{Viva} is defined as the average of multiple QA scores:
$R_{\mathrm{acc}} = \frac{1}{N} \sum_{k=1}^N X_k.$
Using the variance–covariance decomposition, we first write

\begin{small}
\begin{equation}
\mathrm{Var}[R_{\mathrm{acc}}] = \mathrm{Var}\Biggl[\frac{1}{N} \sum_{k=1}^N X_k \Biggr] = \frac{1}{N^2} \mathrm{Var}\Biggl[\sum_{k=1}^N X_k \Biggr].
\end{equation}
\end{small}

Next, applying the formula for the variance of a sum of correlated variables:

\begin{small}
\begin{equation}
\mathrm{Var}\Biggl[\sum_{k=1}^N X_k \Biggr] = \sum_{k=1}^N \mathrm{Var}[X_k] + 2 \sum_{k<l} \mathrm{Cov}[X_k, X_l].
\end{equation}
\end{small}

Combining the two steps, we obtain:

\begin{small}
\begin{equation}
\mathrm{Var}[R_{\mathrm{acc}}] = \frac{1}{N^2} \left( \sum_{k=1}^N \sigma_k^2 + 2 \sum_{k<l} \rho \sqrt{\sigma_k^2 \sigma_l^2} \right).
\end{equation}
\end{small}

\noindent \textbf{Upper Bound on Single QA Variance under Graded Rewards:} For any random variable $X_k \in [0,1]$, the variance satisfies
$\mathrm{Var}[X_k] \le \mu_k (1 - \mu_k),$
with equality if and only if $X_k$ follows a Bernoulli distribution. Since graded rewards allow $X_k$ to take intermediate values, the single QA variance is typically lower than that of binary rewards. Note that in cases where VQA outputs are highly uncertain or clustered near extremes, the variance may approach the upper bound, but it generally does not exceed the binary reward variance.

\noindent \textbf{Implications for Policy Gradient Stability:} In policy gradient methods, a single-step gradient estimate can be written as
$
\hat g = \nabla_\theta \log \pi_\theta(o_i \mid x) \, A_i,
$
where $A_i$ is the advantage estimate constructed from the reward signal. Assuming $\nabla_\theta \log \pi_\theta(o_i \mid x)$ is bounded, the variance of the gradient estimator is positively correlated with the variance of the reward signal. Hence, \textsc{Viva} rewards reduce variance at both the single QA level and the multi-QA aggregation level, suppressing the propagation of reward noise and providing more stable gradient signals for downstream policy optimization. Moreover, the GRPO normalization step further mitigates the impact of extreme rewards on the gradient, enhancing training stability.

\begin{figure*}[t]
\begin{tcolorbox}[
    colback=white,
    colframe=black,
    title=\textbf{Prompts for Topic Generation}
]
\textbf{System Prompt:}\\
You are an AI assistant that specializes in generating structured JSON data.

Your task is to generate a unique, engaging topic for each set of numbered keywords provided by the user.

Output Requirements:\\
1. You MUST return a single, valid JSON object.\\
2. The JSON object MUST contain exactly one key: "topics".\\
3. The value of "topics" MUST be a JSON array of strings.\\
4. Each string in the array should be a 2-3 sentence topic description corresponding to one set of keywords from the user input.\\
5. The number of strings in the array MUST EXACTLY match the number of keyword sets provided.

\vspace{0.3cm}
\textbf{User Prompt:}\\
Please generate topic descriptions for the following 3 characters. Return the result as a JSON object according to the system instructions.

1. Keywords: Name=Alex, Age=32, Profession=Software Engineer, Trait=innovative problem solving, Goal=to streamline a complex workflow\\
2. Keywords: Name=Jordan, Age=45, Profession=Product Manager, Trait=user-centric design, Goal=to map out a new user experience\\
3. Keywords: Name=Taylor, Age=28, Profession=Data Analyst, Trait=data-driven, Goal=to present data insights to stakeholders

\end{tcolorbox}
\caption{The prompt template used for generating different topics to set scene limitations.}
\label{fig:prompt_topic}
\end{figure*}

\begin{figure*}[t]
\begin{tcolorbox}[
    colback=white,
    colframe=black,
    title=\textbf{Prompts for Scenario Generation},
    boxrule=1pt,
    arc=0pt
]
\textbf{System Prompt:}\\
You are an expert in diagram design and have a broad knowledge of different topics.

\textbf{User Prompt:}\\
My scenario is: ``\texttt{\{scenario\}}''\\
I want you to generate \texttt{\{num\_topics\}} topics for a Mermaid ``\texttt{\{figure\_type\}}'' that I will be interested in or I may see during my daily life given my scenario.

Here are the requirements:
\begin{enumerate}
    \item Each topic is a high-level summary of the contents in the diagram with some design details, e.g., ``a Sequence Diagram illustrating a user login process with two-factor authentication''.
    \item The topics should be diverse to help me generate varied diagrams. Each topic should be unique and not overlap with others.
    \item The topics are strictly conditioned on the Mermaid diagram type. Please ensure the topics you provided can be best visualized in ``\texttt{\{figure\_type\}}''.
    \item All topics must be in English, even if the scenario is non-English.
    \item List \texttt{\{num\_topics\}} topics for ``\texttt{\{scenario\}}'' and separate them with a | character, e.g., topic1 | topic2 | ...... | topic\texttt{\{num\_topics\}}.
\end{enumerate}
Do not include any additional text at the beginning or end of your response.

\end{tcolorbox}
\caption{The prompt template used for generating diverse diagram scenario based on user topics and specific diagram types.}
\label{fig:prompt_scenario}
\end{figure*}

\begin{figure*}[t]
\begin{tcolorbox}[
    colback=white,
    colframe=black,
    title=\textbf{Prompts for Structured Data Generation},
    boxrule=1pt,
    arc=0pt
]
\textbf{System Prompt:}\\
You are an expert in diagram design and have broad knowledge about various topics.

\vspace{0.3cm}
\textbf{User Prompt:}\\
My topic is: ``\texttt{\{topic\}}''\\
I need structured data elements about ``\texttt{\{topic\}}'', which can be used to generate a Mermaid ``\texttt{\{figure\_type\}}''.

Here are the requirements:
\begin{enumerate}
    \item The data should be highly structured as a JSON object, with its schema tailored specifically for the ``\texttt{\{figure\_type\}}'' syntax. For example, for a Class Diagram, the JSON should contain a list of `classes' and a list of `relationships'.
    \item The data should be realistic, and the contents should be named using real-world entities. Do not use placeholder names like xxA, xxB, etc.
    \item The elements should be concise and directly map to a meaningful diagram. Do not provide too many elements; just the key information.
    \item All elements must be in English, even if the topic is non-English.
    \item You can use the provided JSON templates to structure decision-based flows. If the topic is related to decision-making or conditional logic, please use or adapt the templates provided in the \texttt{<templates>} block.
\end{enumerate}

\end{tcolorbox}
\caption{The prompt template used for generating structured JSON data elements tailored to specific Mermaid diagram types and topics.}
\label{fig:prompt_data}
\end{figure*}

\begin{figure*}[t]
\begin{tcolorbox}[
    colback=white,
    colframe=black,
    title=\textbf{Prompts for Mermaid Mindmap Code Generation},
    boxrule=1pt,
    arc=0pt
]
\footnotesize 

\textbf{System Prompt:}\\
You are an expert in diagram design and good at writing Mermaid code to generate high-quality, visually appealing diagrams.

\par\noindent\rule{\textwidth}{0.4pt} 

\textbf{User Prompt:}\\
My topic is: ``\texttt{\{topic\}}''. I have a JSON object of structured data about ``\texttt{\{scenario\}}'' which needs to be converted into a Mermaid ``mindmap''.

Here is the data (JSON format): \texttt{<data> \{data\} </data>}

Please write the complete Mermaid code to generate the diagram. Here are the requirements:
\begin{enumerate}[noitemsep, topsep=0pt, leftmargin=*] 
    \item The code must be a valid and complete Mermaid mindmap script adhering strictly to syntax rules.
    \item The code must start with ``mindmap''. The ``mindmap'' keyword is reserved.
    \item \textbf{Indentation is the ONLY way to define the hierarchy}. All child nodes MUST be indented more deeply than their parent. Consistent indentation (e.g., 2 or 4 spaces) is required.
    \item Do not include any additional text outside of the Mermaid code block.
    \item The code must be self-contained within the \texttt{```mermaid} ... \texttt{```} block.
\end{enumerate}

\textbf{CRITICAL RULE: A mindmap can only have ONE SINGLE ROOT NODE.}
\begin{itemize}[noitemsep, topsep=0pt, leftmargin=*] 
    \item The first line of the mindmap defines the root.
    \item All other nodes MUST BE indented under this root node.
    \item Any line with zero indentation (nodes or styles) will be treated as a second root, causing a fatal error.
\end{itemize}

\textbf{IMPORTANT: Styling Syntax and Placement}\\
The safest way is to use \texttt{classDef} at the end and apply it via \texttt{:::}. \textbf{AVOID} placing \texttt{style} commands at the top level.

\textbf{Negative Example (WRONG):}\\
\texttt{mindmap}\\
\texttt{\ \ Root}\\
\texttt{Another Root // <-- FATAL ERROR. Not indented.}

\textbf{Positive Example (Valid Syntax):}\\
\texttt{mindmap}\\
\texttt{\ \ root((Main Topic)):::mainStyle}\\
\texttt{\ \ \ \ Sub-Topic 1}\\
\texttt{\ \ \ \ \ \ Detail A}\\
\texttt{\ \ \ \ \ \ Detail B:::detailStyle}\\
\texttt{\%\% Styles defined at the end}\\
\texttt{classDef mainStyle fill:\#f0f8ff,stroke:\#333,stroke-width:2px}\\
\texttt{classDef detailStyle fill:\#lightgrey,stroke:\#333}

\end{tcolorbox}
\caption{The prompt template used and applying structured JSON data into executable Mermaid Mindmap code.}
\label{fig:prompt_code}
\end{figure*}

\section{Efficiency and Computational Cost}
\label{app:efficiency_analysis}
The offline question generation for 25k samples was completed in approximately 1 hour using the GPT-4o-mini API with a concurrency of 10. All training was conducted on a cluster of 24 NVIDIA H800 GPUs. For the baseline configuration (w/o Reasoning), the SFT and RL stages require 12 and 20 hours for the 3B model, and approximately 16 and 26 hours for the 7B model, respectively. In contrast, the reasoning-enriched (w/ Reasoning) 3B model requires 16 hours for SFT and 30 hours for RL. During RL training, the Qwen3-VL-32B reward model is deployed in FP16 precision, occupying approximately 64 GB of VRAM per GPU. Notably, this reward model exhibits high inference efficiency, and its response latency is fully accounted for within the reported RL training durations. Regarding inference efficiency of the final model, Direct Mode maintains standard LLM latency, while Think Mode introduces a latency increase proportional to the reasoning trajectory length.

\section{Evaluation}
\subsection{Prompt Used in Evaluation}
\label{app:judge}
To ensure reproducibility, we provide the exact system prompts used for our GPT-4.1-based evaluation. 
Figure~\ref{fig:prompt_dia2code} illustrates the prompts for \textbf{Diagram-to-Code}, focusing on Visual Fidelity ($S_{vis}$).
Finally, Figure~\ref{fig:prompt_editing} presents the prompts for \textbf{Diagram Editing}, covering Content Preservation ($S_{pres}$) and Instruction Adherence ($S_{task}$).
Figure~\ref{fig:prompt_text2code} displays the prompts for \textbf{Text-to-Code} evaluation, assessing Visual Correctness ($S_{vis}$) and Task Adherence ($S_{task}$).

\begin{figure*}[t]
\begin{tcolorbox}[
    colback=white,
    colframe=pink,
    title=\textbf{Prompts for \textbf{Diagram-to-code} Evaluation}
]
\textbf{JUDGE\_INSTRUCT\_VISUAL\_FIDELITY:}\\
You are an excellent judge at evaluating the visual fidelity of a diagram reconstruction task. You will be giving scores on how faithfully the reconstructed diagram matches the original source image in terms of both structural logic and visual appearance.\\
The original diagram (source image) will be given to you as the first image.\\
The reconstructed diagram (rendered from model-generated code) will be given to you as the second image.\\
Please score how well the reconstructed diagram matches the original image on a scale from 0 to 100. \\
Scoring should be carried out based on the Visual Fidelity, which considers the following aspects: \\
First, determine the Topological Integrity by strictly evaluating the logical correctness, including the existence of all nodes, the accuracy of connections (edges), and the correctness of the text content (OCR). 
Second, evaluate the Visual Consistency by checking the preservation of visual attributes such as layout orientation (e.g., top-down vs. left-right), node shapes, line styles, and overall aesthetic alignment with the original image.\\
A score of 100 implies that the reconstructed diagram is a perfect digital twin of the original. Please penalize any missing information, incorrect logic, or significant stylistic deviations. \\
After scoring from the above aspects, please give a final score. Do not write anything else. The final score is preceded by the [FINAL SCORE] token.\\
For example [FINAL SCORE]: 45\\

\end{tcolorbox}
\caption{The prompt used for the Diagram-to-Code task evaluation.}
\label{fig:prompt_dia2code}
\end{figure*}

\begin{figure*}[t]
\begin{tcolorbox}[
    colback=white,
    colframe=green,
    title=\textbf{Prompts for \textbf{Diagram Editing} Evaluation}
]
\textbf{JUDGE\_INSTRUCT\_ADHERENCE:}\\
You are an excellent judge at evaluating the instruction adherence of a diagram editing task. You will be giving scores on how well the modified diagram executes the user's request.\\
The original diagram (before editing) will be given to you as the first image.\\
The modified diagram (after editing) will be given to you as the second image.\\
The user instruction describing the required change is provided below (begins from [EDIT INSTRUCTION] token).\\
Please score how well the modified diagram matches the instruction. Score it on a scale from 0 to 100.\\
Scoring should be carried out in the following aspect:\\
Instruction Adherence:\\
Determine strictly if the specific changes requested in the instruction were applied in the modified diagram.\\
Focus ONLY on the requested change (e.g., color change, node addition).
Ignore side effects (e.g., layout shifts) as long as the requested change is present.\\
After scoring from the above aspect, please give a final score. Do not write anything else. The final score is preceded by the [FINAL SCORE] token.\\
For example [FINAL SCORE]: 40\\

\vspace{0.3cm}
\textbf{JUDGE\_INSTRUCT\_PRESERVATION:}\\
You are an excellent judge at evaluating the stability and preservation of a diagram editing task. You will be giving scores on how well the agent preserved the parts of the diagram that should not have changed.
The original diagram (before editing) will be given to you as the first image.\\
The modified diagram (after editing) will be given to you as the second image.\\
The user instruction is provided below to help you identify what should have changed.\\
Please score how well the modified diagram preserves the unrequested content. Score it on a scale from 0 to 100.\\
Scoring should be carried out in the following aspects:
Content Preservation:\\
Compare the nodes, text, and connections that were NOT mentioned in the instruction.\\
Check for layout stability and ensure the agent did not hallucinate new, unrequested nodes or break the visual structure.\\
A score of 100 implies perfect preservation of unrequested areas.\\
After scoring from the above aspect, please give a final score. Do not write anything else. The final score is preceded by the [FINAL SCORE] token.\\
For example [FINAL SCORE]: 30\\

\end{tcolorbox}
\caption{The prompt used for the Diagram Editing task evaluation.}
\label{fig:prompt_editing}
\end{figure*}

\begin{figure*}[t]
\begin{tcolorbox}[
    colback=white,
    colframe=blue,
    title=\textbf{Prompts for \textbf{Text-to-Code} Evaluation}
]
\textbf{JUDGE\_INSTRUCT\_VIS:}\\
You are an excellent judge at evaluating visualization plots between a model generated plot and the ground truth. You will be giving scores on how well it matches the ground truth plot.\\
The generated plot will be given to you as the first figure.\\
Another plot will be given to you as the second figure, which is the desired outcome of the user query, meaning it is the ground truth for you to reference.
Please compare the two figures head-to-head and rate them.\\
Suppose the second figure has a score of 100, rate the first figure on a scale from 0 to 100.\\
Scoring should be carried out in the following aspect:\\
Plot correctness:\\
Compare closely between the generated plot and the ground truth...\\
After scoring from the above aspect, please give a final score. Do not write anything else. The final score is preceded by the [FINAL SCORE] token.\\
For example [FINAL SCORE]: 40\\

\vspace{0.3cm}
\textbf{JUDGE\_INSTRUCT\_TASK:}\\
You are an excellent judge at evaluating a visualization plot according to the given task. You will be giving scores on how well the plot image matches the task.\\
The generated plot will be given to you as an image.\\
Please score how well the plot matches the task. Score it on a scale from 0 to 100.
Scoring should be carried out in the following aspect:\\
Task adherence: how the plot corresponds to the task given below (begins from [PLOT TASK] token)\\
After scoring from the above aspect, please give a final score. Do not write anything else. The final score is preceded by the [FINAL SCORE] token. \\
For example [FINAL SCORE]: 40\\

\end{tcolorbox}
\caption{The prompt  used for Text-to-Code task evaluation.}
\label{fig:prompt_text2code}
\end{figure*}

\subsection{Evaluation Metrics}
\label{app:metrics}

We utilize a GPT-4.1-based evaluation pipeline to simulate human judgment across diverse tasks, supplemented by deterministic code-level metrics. All scores are normalized to a scale of 0 to 100.

\subsubsection{Diagram-to-Code Metrics}
For the reverse engineering task, we decouple the structural rendering from the underlying code logic:

\noindent \textbf{Visual Fidelity ($S_{vis}$):} A composite metric evaluating the preservation of visual attributes (e.g., node shapes, layout orientation) and topological integrity (e.g., nodes, edges, and OCR content) between the original image and the reconstructed rendering.
\noindent \textbf{Code Accuracy ($S_{code}$):} Evaluates the syntactic and logical similarity of the generated code against the ground truth using \textit{CrystalBLEU}~\cite{eghbali2022crystalbleu}, which mitigates the bias of common code templates.

\subsubsection{Diagram Editing Metrics}
This task focuses on the precision of targeted modifications and the stability of unchanged components:

\noindent \textbf{Instruction Adherence ($S_{task}$):} Measures the success rate of specific modifications requested by the user (e.g., color changes or node deletions), focusing strictly on the execution of the edit instruction.
\noindent \textbf{Content Preservation ($S_{pres}$):} Assesses the stability of regions not targeted by the edit. It penalizes structural collapse, hallucinations, or unintended semantic drifts in unmodified parts of the diagram.

\subsubsection{Text-to-Code Metrics}
Following the methodology established in \textit{VisCoder2}~\cite{ni2025viscoder2}, we evaluate the synthesis quality across two dimensions:

\noindent \textbf{Visual Correctness ($S_{vis}$):} Evaluates the alignment between the generated diagram's visual appearance and the ground truth, focusing on data distribution and geometric relationships.
\noindent \textbf{Task Adherence ($S_{task}$):} Measures how accurately the model follows textual constraints (e.g., specific data points or plot types), independent of the final rendering's aesthetic quality.

\section{Inference Result of Omni Diagram}

\subsection{Examples of Ability}
\label{examples_of_ability}

To further substantiate the quantitative results presented in the main text, we provide a comprehensive gallery of qualitative examples in Figure~\ref{fig:viva_showcase}. 
This visualization demonstrates the versatility and robustness of our model across the three target diagrammatic languages: \LaTeX{}, Mermaid, and PlantUML. 
The figure is stratified into three columns, each corresponding to a core task:

\noindent \textbf{Diagram-to-Code:} The left column compares the input ground truth images with the model's reconstructed outputs. The results exhibit exceptional visual fidelity and topological correctness, accurately recovering complex structures such as molecular energy levels, sequence interactions, and class hierarchies.

\noindent \textbf{Diagram Editing:} The middle column illustrates the model's capability to perform precise, localized modifications based on user instructions. As highlighted by the red bounding boxes, the model successfully executes specific edits (e.g., text updates, node recoloring) while preserving the integrity of the unmodified regions.
    
\noindent \textbf{Text-to-Code:} The right column showcases the model's ability to synthesize complex diagrams from dense, long-context natural language descriptions. The examples cover diverse scenarios, including system architectures, Kanban boards, and sensor networks, confirming the model's strong instruction-following capabilities in zero-shot generation.

\subsection{Failure Mode Analysis}
\label{app:failure_analysis}

The error distribution across different model scales reveals clear boundaries in current diagram generation capabilities, and the specific error statistics are summarized in Table \ref{tab:error_stats}. For LaTeX tasks, reference errors represent the most frequent failure mode because the model often struggles to maintain consistency in diagrams with dense connections and complex coordinate systems. The transition from the 3B to the 7B model shows a significant reduction in these instances, which suggests that larger models possess a better capacity for the spatial memory required to link node identifiers across extensive code.

Conversely, syntax errors in PlantUML and Mermaid remain largely stagnant regardless of model size, pointing to a persistent bottleneck in handling strict domain specific languages. These failures typically occur in diagrams featuring multi-layer nesting or highly recursive structures where the model fails to properly manage hierarchical blocks. When faced with extremely dense wiring or deep logic tiers, the output frequently violates language constraints, demonstrating that topological complexity remains a primary challenge for ensuring structural integrity during the generation process.

\begin{table}[h]
  \centering
  \resizebox{\columnwidth}{!}{%
    \begin{tabular}{llcc}
      \toprule
      \multirow{2}{*}{\textbf{Language}} & \multirow{2}{*}{\textbf{Error Category}} & \multicolumn{2}{c}{\textbf{OmniDiagram (RL)}} \\ 
      
      \cmidrule(lr){3-4} 
      
       & & \textbf{3B} & \textbf{7B} \\ 
      
      \midrule
      
      \multirow{6}{*}{\textbf{LaTeX}} 
        & Reference Error & 33 & 23 \\
        & Structural Error & 16 & 10 \\
        & Syntax Error & 15 & 13 \\
        & Math \& Resource Error & 8 & 9 \\
        & Dependency \& Encoding & 7 & 6 \\
        & Timeout & 0 & 2 \\
      \midrule
      
      \multirow{3}{*}{\textbf{PlantUML}} 
        & Syntax Error & 83 & 83 \\
        & Logic Error & 3 & 6 \\
        & System Error & 0 & 2 \\
      \midrule
      
      \multirow{5}{*}{\textbf{Mermaid}} 
        & Syntax Error & 52 & 52 \\
        & Rendering \& Resource Error & 6 & 6 \\
        & Logic Error & 5 & 5 \\
        & Structural Error & 3 & 4 \\
        & Formatting Error & 2 & 0 \\
        
      \bottomrule
    \end{tabular}%
  } 
\caption{Statistical summary of generation rrrors across diagram languages.}
\label{tab:error_stats}
\vspace{-13pt}
\end{table}

\section{Detailed Analysis of Reasoning Strategies}

\label{app:ablation_analysis}

In this section, we provide a granular analysis of the trade-offs between reasoning-enhanced training and inference strategies, as presented in Table~\ref{tab:ablation_thinking_rl_detailed}.

\noindent \textbf{Internalization vs. Explicit Output.}
Our results demonstrate that the benefits of Chain-of-Thought data are best realized through \textit{internalization} rather than explicit output. The \textit{Mixed SFT} model in Direct mode achieves a higher Task Adherence score ($S_{task}$ 63.0) compared to the \textit{Pure SFT} baseline (61.0), confirming that the model learns to plan better diagrams from the reasoning data. However, explicitly outputting these thoughts acts as a disturbance during inference.

\noindent \textbf{The "Syntax Tax".}
A critical discovery is the trade-off between semantic quality and syntactic robustness, which we term the "Syntax Tax." 
When valid, diagrams generated with explicit thinking exhibit superior visual details (Valid $S_{vis}$ \underline{59.23} vs. 55.32). 
However, the generation of extensive natural language reasoning consumes the context window and disrupts the model's ability to maintain the strict syntactic structure required for diagrammatic code (e.g., \LaTeX{} or Mermaid). 
This results in a precipitous drop in execution rate (from 84.73\% to 69.47\%), causing the overall performance to suffer despite higher potential visual quality.

\noindent \textbf{Reinforcement Learning alignment.}
RL effectively mitigates robustness issues but does not alter the fundamental superiority of direct inference for this task. 
RL aligns the model with the compiler's requirements, boosting the execution rate to a state-of-the-art \textbf{90.84\%} in Direct mode. 
While RL also improves the robustness of the Thinking mode, the structural instability of mixing long-form text with code remains, making Direct inference the optimal strategy for Diagram-to-Code generation.

\begin{table}[htbp]
\centering
\setlength{\tabcolsep}{3pt} 
\resizebox{\columnwidth}{!}{
\begin{tabular}{ll c cc cc}
\toprule
\multirow{2}{*}{\textbf{Model Setting}} & \multirow{2}{*}{\textbf{Inf.}} & \textbf{Rob.} & \multicolumn{2}{c}{\textbf{Rendered (v)}} & \multicolumn{2}{c}{\textbf{Overall}} \\
\cmidrule(lr){3-3} \cmidrule(lr){4-5} \cmidrule(lr){6-7}
 & & Exec (\%) & $S_{\text{vis}}$ & $S_{\text{task}}$ & $\textbf{S}_{\text{vis}}$ & $\textbf{S}_{\text{task}}$ \\
\midrule
Qwen2.5-VL-3B & Direct & 51.2 & 43.7 & 62.3 & 22.0 & 32.0 \\
VisCoder2-3B & Direct & 76.3 & -- & -- & 43.0 & 59.0 \\
\midrule
3B\_SFT (190k) & Direct & 82.4 & 55.1 & 77.9 & 45.8 & 64.3 \\
\multirow{2}{*}{3B\_SFT (270k)} & Direct & 84.7 & 55.3 & 73.9 & \textbf{47.0} & \textbf{63.0} \\
 & Think & 69.5 & \underline{59.2} & \underline{75.0} & 41.0 & 52.0 \\
\midrule
3B\_SFT+RL (190k) & Direct & 88.6 & 55.8 & 72.8 & 49.4 & 64.5 \\
\multirow{2}{*}{3B\_SFT+RL(270k)} & Direct & \textbf{90.8} & 54.2 & 76.6 & \textbf{49.0} & \textbf{70.0} \\
 & Think & 75.6 & 57.3 & \textbf{76.7} & 43.0 & 58.0 \\
\bottomrule
\end{tabular}
}
\caption{Ablation study on data composition and inference strategies. \textbf{(v)} metrics are computed exclusively on successfully rendered images. \underline{Underlined} values highlight superior visual quality. Overall score $S \approx \text{Exec} \times \text{Rendered}$.}
\label{tab:ablation_thinking_rl_detailed}
\end{table}

\end{document}